\documentclass[11pt]{article}       
\usepackage{geometry}               
\usepackage{graphicx}
\usepackage{caption}
\usepackage{subcaption}
\usepackage{comment}
\usepackage{amsmath} 
\usepackage{amssymb}  
\usepackage{amsthm}  
\usepackage{bm}
\usepackage{lipsum}
\usepackage[linesnumbered,ruled,vlined]{algorithm2e}
\usepackage{color}
\usepackage{array}
\usepackage{hyperref}
\usepackage{upgreek}
\usepackage{todonotes}
\usepackage{cite}

\usepackage{multirow}

\newtheorem{proposition}{Proposition}

\newtheorem{lemma}{Lemma}

\numberwithin{equation}{section}

\newcommand{\dist}{\mathrm{dist}}

\newcommand{\cvar}{\mathrm{CVaR}}

\title{Wasserstein Distributionally Robust Motion Control for Collision Avoidance Using Conditional Value-at-Risk} 

\author{Astghik~Hakobyan
\and
 Insoon Yang\thanks{Department of Electrical and Computer Engineering, Automation and Systems Research Institute,  Seoul National University, Seoul 08826, Korea, (\{astghikhakobyan, insoonyang\}@snu.ac.kr). This work was supported in part by NSF under ECCS-1708906, in part by 
the Creative-Pioneering Researchers Program through SNU, the Basic Research Lab Program through the National Research Foundation of Korea funded by the MSIT(2018R1A4A1059976), and Samsung Electronics. }
}

\date{}

\begin{document}
\maketitle
\pagestyle{myheadings}
\thispagestyle{plain}

\begin{abstract}
In this paper, a risk-aware motion control scheme is considered for mobile robots to avoid randomly moving obstacles when the true probability distribution of uncertainty is unknown. 
We propose a novel model predictive control (MPC) method for limiting the risk of unsafety even when the true distribution of the obstacles' movements deviates, within an \emph{ambiguity set}, from the empirical distribution obtained using a limited amount of sample data. 
By choosing the ambiguity set as a statistical ball with its radius measured by the \emph{Wasserstein metric},
we achieve a probabilistic guarantee of the
\emph{out-of-sample risk}, evaluated using new sample data generated independently of the training data.
To resolve the infinite-dimensionality issue inherent in the distributionally robust MPC problem, 
we reformulate it as a finite-dimensional nonlinear program
using modern distributionally robust optimization techniques based on the Kantorovich duality principle. 
To find a globally optimal solution in the case of affine dynamics and output equations, a spatial branch-and-bound algorithm is designed using McCormick relaxation.
The performance of the proposed method is demonstrated and analyzed through simulation studies using a nonlinear car-like vehicle model and a linearized quadrotor model.
\end{abstract}

\section{Introduction}

Safety is one of the most fundamental challenges in the operation of
mobile robots and autonomous vehicles in practical environments, which are uncertain and dynamic. 
In particular, the unexpected movement of objects and agents often jeopardizes the collision-free navigation of mobile robots. 
Unfortunately, predicting an object's motion is a challenging task in many circumstances due to the lack of knowledge about the object's possibly uncertain dynamics. 
Estimating an accurate probability distribution of underlying uncertainty
often demands large-scale high-resolution sensor measurements over a long training period.
The research question to be addressed in this work is as follows: \emph{Can a robot make a safe decision using an unreliable distribution estimated from small samples?}
To answer this question, we develop an optimization-based motion control method that uses a limited amount of data for making a risk-aware decision with a finite-sample probabilistic guarantee of collision avoidance.

Several risk-sensitive decision-making methods have been proposed for robots to avoid obstacles in uncertain environments. 
Chance-constrained methods are among the most popular approaches, as they can be used to directly limit the probability of collision. 
Because of their intuitive and practical role, chance constraints have been extensively used in sampling-based planning~\cite{Luders2010, Bry2011, Summers2018} and model predictive control (MPC)~\cite{Blackmore2011, DuToit2012}.
However, it is computationally challenging to handle a chance constraint due to its nonconvexity. 
This often limits the admissible class of probability distributions and system dynamics and/or requires an undesirable approximation.
To resolve the issue of nonconvexity, a few theoretical and algorithmic tools have been developed using a particle-based approximation~\cite{Blackmore2010} and
 semidefinite programming formulation~\cite{Jasour2015}, among others.
 Another approach is to use a convex risk measure, which is computationally tractable. 
In particular, \emph{conditional value-at-risk} (CVaR) has recently drawn a great deal of interest in motion planning and control~\cite{Chow2015, Majumdar2017isrr, Singh2018, Hakobyan2019}.
The CVaR of a random loss represents the conditional expectation of the loss within the $(1 - \alpha)$ worst-case quantile of the loss distribution, where $\alpha \in (0, 1)$~\cite{Rockafellar2002a}.
As claimed in \cite{Majumdar2017isrr}, CVaR is suitable for rational risk assessments in robotic applications because of its \emph{coherence} in the sense of Artzner {\it et al.}~\cite{Artzner1999}. 
In addition to its computational tractability, CVaR is capable of distinguishing the worst-case tail events, and thus it is effective to take into account rare but unsafe events.
To enjoy these advantages,
we adopt CVaR to measure the risk of unsafety.

The performance of such risk-aware motion control tools critically depends on the quality of information about the probability distribution of underlying uncertainties, such as an obstacle's random motion. 
If a poorly estimated distribution is used, 
it may cause unwanted behaviors of the robot, leading to a collision. 
One of the most straightforward ways to estimate the probability distribution is to collect the sample data of an obstacle's movement and construct an empirical distribution.
The use of an empirical distribution is equivalent to a sample average approximation (SAA) of the stochastic programs~\cite{Shapiro2014}. 
Although SAA is quite effective with asymptotic optimality, 
it does not have a finite-sample guarantee of satisfying risk constraints.
In our previous work using SAA, it was empirically observed that risk constraints are likely to be violated when the sample size is very small~\cite{Hakobyan2019}.

To account for this issue of limited distributional information, 
we seek an efficient risk-aware motion control method that is robust against distribution errors. 
Our method is based on \emph{distributionally robust optimization} (DRO), which is employed to solve a stochastic program in the face of the worst-case distribution drawn from a given set, called the \emph{ambiguity set}~\cite{Calafiore2006, Delage2010, Wiesemann2014}.
In this work, we use the Wasserstein ambiguity set, a statistical ball that contains all the probability distributions whose \emph{Wasserstein distance} from an empirical distribution is no greater than a certain radius~\cite{Esfahani2018, Zhao2018, Gao2016}.
The Wasserstein ambiguity set has several salient features, such as providing a non-asymptotic performance guarantee and addressing the closeness between two points in the support, unlike other statistical distance-based ambiguity sets (e.g., using phi-divergence)~\cite{Gao2016, BenTal2013, Bayraksan2015}.
The proposed motion control method is robust against obstacle movement distribution errors characterized by the Wasserstein ambiguity set.

The contributions of this work can be summarized as follows.
First, a novel model predictive control (MPC) method is proposed to limit the risk of unsafety through CVaR constraints that must hold for any perturbation of the empirical distribution within the Wasserstein ambiguity set. 
Thus, the resulting control decision is guaranteed to satisfy the risk constraints for avoiding randomly moving obstacles in the presence of allowable distribution errors.
Moreover, the proposed method provides a finite-sample probabilistic guarantee of limiting \emph{out-of-sample risk}, meaning that the risk constraints are satisfied with probability no less than a certain threshold even when evaluated with new sample data chosen independently of the training data.
Second, for computational tractability, we reformulate the distributionally robust MPC (DR-MPC) problem, which is infinite-dimensional, into a finite-dimensional nonconvex optimization problem.
The proposed reformulation procedure is developed using modern DRO techniques based on the Kantorovich duality principle~\cite{Esfahani2018}.
Third, a spatial branch-and-bound (sBB) algorithm is designed with McCormick relaxation to address the issue of nonconvexity.
The proposed algorithm finds a globally optimal control action in the case of affine system dynamics and output equations. 
The performance and utility of the proposed method are demonstrated through two simulation studies, one with a nonlinear car-like vehicle model and another with a linearized quadrotor model.
The results of numerical experiments confirm that, even when the sample size is small, the proposed DR-MPC method can successfully avoid randomly moving obstacles with a guarantee of limiting out-of-sample risk, while its SAA counterpart fails to do so. 
%

The rest of this paper is organized as follows.
In Section~\ref{sec:form}, the problem setup is introduced and the Wasserstein DR-MPC problem is formulated using CVaR constraints for collision avoidance. 
In Section~\ref{sec:reform}, a set of reformulation procedures is proposed to resolve the infinite-dimensionality issue inherent in the DR-MPC problem. 
Section~\ref{sec:sBB} is devoted to a spatial branch-and-bound algorithm for solving the reformulated optimization problem, which is nonconvex. 
In Section~\ref{sec:outofsample}, the probabilistic guarantee of limiting out-of-sample risk is discussed using the measure concentration inequality for Wasserstein ambiguity sets. 
Finally, the simulation results are presented and analyzed in Section~\ref{sec:sim}.

\section{Problem Formulation}\label{sec:form}
\subsection{System and Obstacle Models}

In this paper, we consider a mobile robot, which can be modeled by the following discrete-time dynamical system:
\begin{equation}\nonumber
\begin{split}
x(t+1)&=f(x(t),u(t))\\
y(t)&=h(x(t),u(t)),
\end{split}
\end{equation}
where $x(t)\in \mathbb{R}^{n_x}$, $u(t)\in\mathbb{R}^{n_u}$ and $y(t)\in\mathbb{R}^{n_y}$ are the system state, the control input, and the system output, respectively. 
In general, $f: \mathbb{R}^{n_x} \times \mathbb{R}^{n_u} \to \mathbb{R}^{n_x}$ and $h: \mathbb{R}^{n_x} \times \mathbb{R}^{n_u} \to \mathbb{R}^{n_y}$ are nonlinear functions, representing the system dynamics and the output mapping, respectively.
We regard the output as the robot's current position in the $n_y$-dimensional configuration space. 
Typical robotic systems operate under some state and control constraints:
\[
x(t)\in \mathcal{X},\quad u(t)\in\mathcal{U}.
\]
We assume that $\mathcal{X}\subseteq\mathbb{R}^{n_x}$ and $\mathcal{U}\subseteq\mathbb{R}^{n_u}$ are convex sets.

\begin{figure}[tb]
\centering
\includegraphics[width=5in]{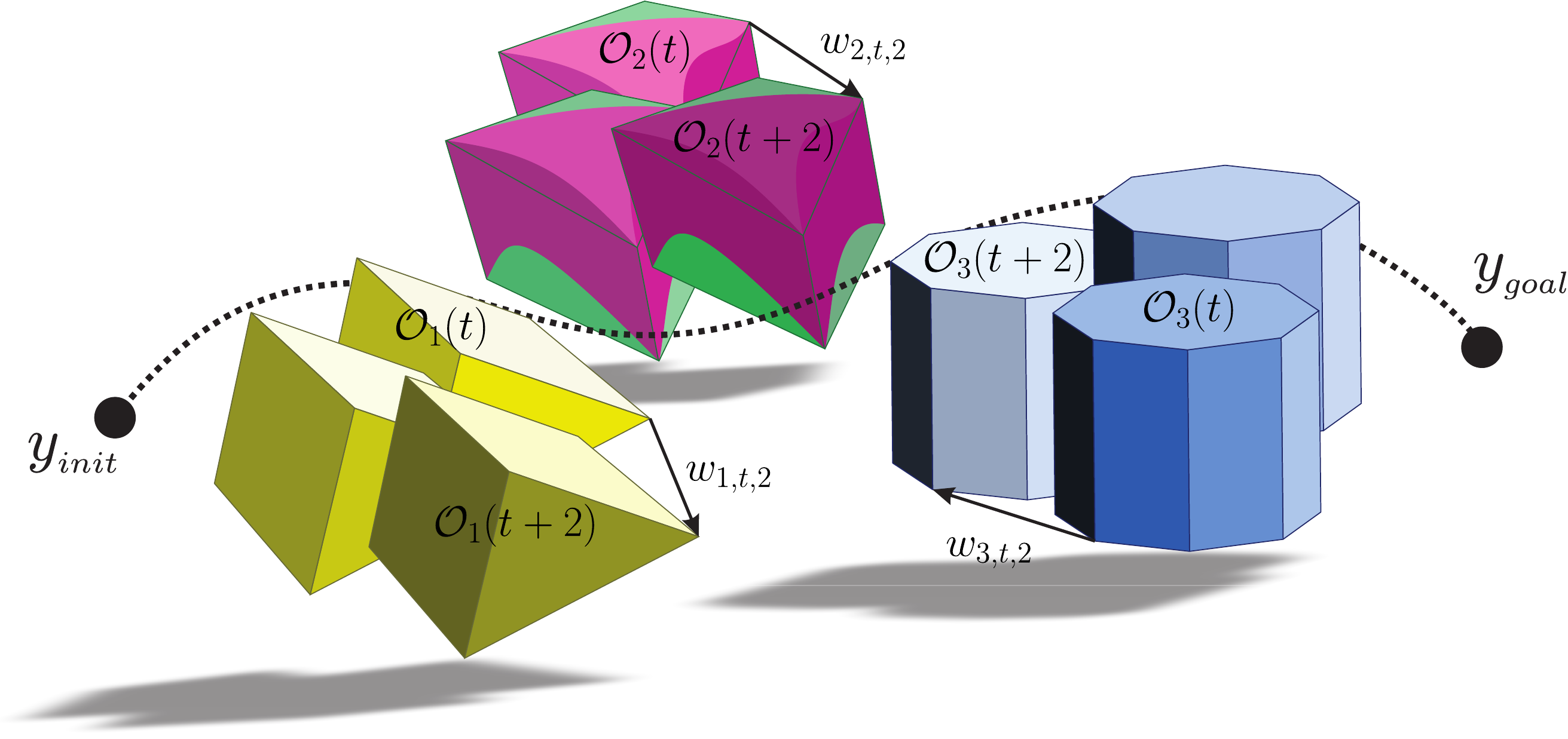}
\caption{Robot configuration space with randomly moving obstacles.}
\label{fig:obs}
\end{figure}

To formulate a collision avoidance problem, we consider \emph{L randomly moving} rigid body obstacles that the robotic vehicle has to avoid while navigating the configuration space. 
Let the region occupied by the obstacle $\ell$ at stage $t$ be denoted by $\mathcal{O}_\ell(t)\subset\mathbb{R}^{n_y}$. If $\mathcal{O}_\ell (t)$ is not a convex polytope, we over-approximate it as a polytope and choose its convex hull as illustrated in the second obstacle in Fig.~\ref{fig:obs}. 
Each obstacle's motion between two stages is assumed to be modeled by translation:\footnote{Our method may also handle the rotation of obstacles by using the model proposed in our previous work~\cite{Hakobyan2019}. However, for ease of exposition, we only consider translational motion.}
\[
\mathcal{O}_\ell(t+k)=\mathcal{O}_\ell(t)+w_{\ell,t,k},
\]
where $w_{\ell,t,k}$ is a random translation vector in $\mathbb{R}^{n_y}$. An example of obstacles' movements is illustrated in Fig.~\ref{fig:obs}. Here, the sum of a set $\mathcal{A}$ and a vector $w$ is defined by adding $w$ to all elements of $\mathcal{A}$, i.e. $\mathcal{A}+w:=\{a+w\mid a\in \mathcal{A}\}$.

Regarding obstacle $\ell$, we define the \emph{safe region}, denoted by $\mathcal{Y}_\ell (t)$, as the complement to the region occupied by the obstacle, i.e.,
\[
\mathcal{Y}_\ell(t):=\mathbb{R}^{n_y}\setminus\mathcal{O}_{\ell}^o(t),
\]
where $\mathcal{O}_\ell^{o}(t)$ denotes the interior of $\mathcal{O}_{\ell}(t)$.
For collision avoidance, it is desirable for the robotic vehicle to navigate in the intersection of safe regions regarding all the obstacles: 
\[
y(t) \in \mathcal{Y}_\ell (t) \quad \forall \ell.
\]
Since obstacle $\ell$ is moving randomly, so is the corresponding safe region. Specifically, it evolves with
\begin{align*}
\mathcal{Y}_\ell(t+k)&=\mathbb{R}^{n_y}\setminus\mathcal{O}_\ell^o(t+k)\\
&=\{x\in\mathbb{R}^{n_y}\mid x\notin \mathcal{O}_\ell^o(t+k)\}\\
&=\{x+w_{\ell,t,k}\in\mathbb{R}^{n_y}\mid x\notin O_{\ell}^o (t)\}\\
&=\mathcal{Y}_\ell(t)+w_{\ell,t,k}.
\end{align*}

\subsection{Reference Trajectory Planning}\label{sec:RRT}

In the offline planning stage, 
a reference trajectory is generated using path-planning tools.
For this work, we employ RRT*~\cite{karaman2011sampling}. 
This particular tool efficiently searches   nonconvex, high-dimensional spaces by randomly building a space-filling tree. The tree is constructed incrementally in a way that quickly reduces the expected distance between a randomly-chosen point and the tree.
It provides an asymptotically optimal solution using tree rewiring and near neighbor search to improve the path quality. The tree starts from an initial state $y_{init}$ and expands to find a path towards the goal state $y_{goal}$, by randomly sampling the configuration space of obstacles in their initial positions and  steering towards the random sample. However, the path generated by RRT* might not be possible to trace, given the dynamics of a robotic vehicle. 
In order to generate a traceable trajectory that takes into account robot dynamics, we perform kinodynamic motion planning based on RRT*~\cite{karaman2010optimal}. 
The major   difference from the baseline RRT* algorithm is that the vehicle dynamics is used for local steering to return a trajectory connecting two states while minimizing the distance between them.

The resulting trajectory is collision-free, given the initial configuration of the obstacles. However, it might not be safe  to follow when the obstacles start to move.
To limit this risk of unsafety during the operation of the robot, 
we propose a more sophisticated motion control tool that takes into account randomly moving obstacles in a distributionally robust manner.

%

\subsection{Measuring Safety Risk Using CVaR}

Our motion control tool uses the notion of \emph{safety risk} introduced in our previous work~\cite{Samuelson2018}. We measure the \emph{loss of safety} regarding obstacle $\ell$ as the deviation of the robot's position from the safe region $\mathcal{Y}_\ell(t)$:
\begin{equation}\label{loss}
\dist(y(t),\mathcal{Y}_\ell(t)):=\min_{a\in\mathcal{Y}_\ell(t)}\|y(t)-a\|_2,
\end{equation}
where $\| \cdot \|_2$ is the standard Euclidean norm.
It is ideal to drive the robot so that the loss of safety is zero. However, in practice, the resulting decision might be overly cautious. Rather than employing such a deterministic approach, we take a stochastic approach by measuring the \emph{safety risk} regarding obstacle $\ell$ as the conditional value-at-risk (CVaR) of safety loss. 
The CVaR of a random loss $X$ is equal to the conditional expectation of the loss within the $(1-\alpha)$ worst-case quantile of the loss distribution and is defined by
\[
\cvar_\alpha(X):=\min_{z\in\mathbb{R}}\mathbb{E} \Bigg[z+\frac{(X-z)^+}{1-\alpha}\Bigg],
\]
where $(\mathbf{x})^+=\max\{\mathbf{x},0\}$.
Accordingly, the safety risk measures the conditional expectation of the distance between the robot position $y(t)$ and the safe region $\mathcal{Y}(t)$ within the $(1-\alpha)$ worst-case quantile of the safety loss distribution.
Our motion control tool seeks the robot's actions that satisfy the following \emph{risk constraint}:
\begin{equation}
\cvar^{\mu_\ell}_\alpha[\dist(y(t),\mathcal{Y}_\ell(t))]\leq\delta_\ell \quad \forall \ell, \label{eq:risk_const}
\end{equation}
where $\delta_\ell \geq 0$ is a user-specified parameter that adjusts risk-tolerance of the robot.

\subsection{Wasserstein Distributionally Robust MPC}

Computing safety risk requires information about the probability distribution of $w_{\ell, t, s}$'s.
 However, the exact probability distribution is unknown in practice, and 
obtaining a reliable distribution is a challenging task. 
In most cases,  we only have a limited amount of sample data generated from the underlying distribution. Probably the simplest way to incorporate the available data into the motion control problem is to employ an empirical distribution as in SAA of stochastic programs~\cite{Shapiro2014}. 
 Specifically, given sample data $\{\hat{w}^{(1)}_{\ell,t,k},\dots,\hat{w}^{(N_k)}_{\ell,t,k}\}$ of $w_{\ell,t,k}$, the empirical distribution is defined as
\begin{equation}
\nu_{\ell,t,k}:=\frac{1}{N_k}\sum_{i=1}^{N_k} \bm{\delta}_{\hat{w}^{(i)}_{\ell,t,k}},\label{emp}
\end{equation}
where $\bm{\delta}_{w}$ is the Dirac delta measure concentrated at $w$. However, this empirical distribution is not capable of reliably estimating the safety risk, particularly when the sample size $N_k$ is small. 
This fundamental limitation results in unsafe decision-making without respecting the original risk constraint. 
Thus, the approach of using empirical distributions may lead to damaging collisions as the safety risk is poorly assessed.

To resolve the issue of unreliable distribution information,
we take a DRO approach.
Instead of using the risk constraint~\eqref{eq:risk_const}, we limit the safety risk evaluated under the worst-case distribution of $w_{\ell, t, k}$ lying in a given set $\mathbb{D}_{\ell,t,k}$, called an \emph{ambiguity set}.
More precisely, we impose the following \emph{distributionally robust risk constraint}:
\[
\sup_{\mu_{\ell,t,k}\in\mathbb{D}_{\ell,t,k}}\mathrm{CVaR}^{\mu_{\ell,t,k}}_\alpha[\mathrm{dist}(y_k,\mathcal{Y}_\ell(t+k))]\leq\delta_\ell \quad \forall \ell.
\]

By limiting the worst-case risk value that the robot can bear, the resulting control action is robust against distribution errors characterized by the ambiguity set. 
In this work, the ambiguity set is chosen as
 the following statistical ball centered at the empirical distribution~\eqref{emp} with radius $\theta > 0$:
\begin{equation}\label{ball}
\mathbb{D}_{\ell, t, k} :=\{\mu \in \mathcal{P} (\mathbb{W}) \mid W (\mu,\nu_{\ell, t, k}) \leq \theta\},
\end{equation}
where $\mathcal{P}(\mathbb{W})$ denotes the set of Borel probability measures on the support $\mathbb{W} \subseteq \mathbb{R}^{n_y}$. Here, the Wasserstein distance (of order 1) $W (\mu,\nu)$ between $\mu$ and $\nu$ 
represents the minimum cost of redistributing mass from one measure to another using a small non-uniform perturbation, and is defined by
\begin{equation} \nonumber
\begin{split}
W (\mu, \nu) := \min_{\kappa \in \mathcal{P}(\mathbb{W}^2)} \Big \{
& \int_{\mathbb{W}^2} \| w - w' \| \; \mathrm{d} \kappa (w, w')  \mid \Pi^1 \kappa = \mu, \Pi^2 \kappa = \nu
\Big \},
\end{split}
\end{equation}
where
$\Pi^i \kappa$ denotes the $i$th marginal of the transportation plan $\kappa$ for $i=1, 2$, and $\| \cdot \|$ is an arbitrary norm on $\mathbb{R}^{n_y}$.
It is worth mentioning that other types of ambiguity sets can be chosen in the proposed DR-MPC formulation.
A popular choice in the literature of DRO is moment-based ambiguity sets~\cite{Calafiore2006, Delage2010, Wiesemann2014}. 
However, such ambiguity sets are often overly conservative and require a large sample size to reliably estimate moment information. 
Statistical distance-based ambiguity sets have also received a considerable interest, by using phi-divergence~\cite{BenTal2013} and Wasserstein distance~\cite{Esfahani2018, Zhao2018, Gao2016, Hota2019}, among others. 
However, unlike other statistical distance-based ones, the Wasserstein ambiguity set contains a richer set of relevant distributions, and the corresponding Wasserstein DRO
 provides a superior finite-sample performance guarantee \cite{Esfahani2018}.
These desirable features play an important role in the proposed motion control tool. 

Finally, we formulate the risk-aware motion control problem as the following \emph{Wasserstein distributionally robust} MPC (DR-MPC) problem:\footnote{Our problem formulation and solution method is different from the one studied by Coulson \emph{et al.}~\cite{Coulson2019} as they consider uncertainties in systems, whereas we consider uncertainties in obstacles' motions. Dynamic programming approaches to distributionally robust optimal control problems have also been studied in~\cite{Xu2012, Yang2017lcss, Yang2017aut, Tzortzis2019}.}
\begin{subequations}\label{DRMPC}
\begin{align}
\inf_{\mathbf{u,x,y}} \; & J(x(t),\mathbf{u}):=\sum_{k=0}^{K-1} r(x_k,u_k)+q(x_K)\label{DRMPCcost}\\
\mathrm{s.t.} \; & x_{k+1}=f(x_k,u_k) \label{DRMPCcons1}\\
&y_k=h(x_k,u_k) \label{DRMPCcons2}\\
&x_0=x(t)\label{CVaRcons3}\\
&x_k\in \mathcal{X}\label{DRMPCcons4a}\\
&u_k\in \mathcal{U}\label{DRMPCcons4b}\\
&\sup_{\mu_{\ell,t,k}\in\mathbb{D}_{\ell,t,k}}\mathrm{CVaR}^{\mu_{\ell,t,k}}_\alpha[\mathrm{dist}(y_k,\mathcal{Y}_\ell(t+k))]\leq\delta_\ell,\label{DRMPCcons5}
\end{align}
\end{subequations}
where $\bold{u} := (u_0, \ldots, u_{K-1})$, $\bold{x} := (x_0, \ldots, x_K)$, $\bold{y}:= (y_0, \ldots, y_{K})$. The constraints \eqref{DRMPCcons1} and \eqref{DRMPCcons4b} should be satisfied for $k=0,\dots,K-1$, the constraints \eqref{DRMPCcons2} and \eqref{DRMPCcons4a} should hold for $k=0,\dots,K$, and the constraint \eqref{DRMPCcons5} is imposed for $k=1,\dots,K$ and $\ell = 1, \ldots, L$. Here, the stage-wise cost function $r: \mathbb{R}^{n_x} \times \mathbb{R}^{n_u} \to \mathbb{R}$ and the terminal cost function $q: \mathbb{R}^{n_x} \to \mathbb{R}$ are chosen to penalize the deviation from the reference trajectory $x^{ref}$ generated in Section \ref{sec:RRT} and to minimize the control effort. Specifically, we set 
\[
J(x(t),\mathbf{u}):=\|x_K-x_K^{ref}\|_P^2+\sum_{k=0}^{K-1}\|x_k-x_k^{ref}\|_Q^2+\|u_k\|^2_R,
\]
 where $Q\succeq 0$, $R\succ 0$ are the weight matrices for state and input, respectively, and $P\succeq 0$ is chosen in a way to ensure stability.
The constraints \eqref{DRMPCcons1} and \eqref{DRMPCcons2} account for the system state and output predicted in the MPC horizon when $x_0$ is initialized as the current state $x(t)$, and \eqref{DRMPCcons4a} and \eqref{DRMPCcons4b} are the constraints on system state and control input, respectively.
The distributionally robust risk constraint is specified in \eqref{DRMPCcons5}, which is the most important part in this problem for safe motion control with limited distribution information. 

The Wasserstein DR-MPC problem is defined and solved in a receding horizon manner. 
Once an optimal solution $\bold{u}^\star$ is obtained given the current state $x(t)$, the first component $u_0^\star$ of $\bold{u}^\star$ is selected as the control input at stage $t$, i.e., $u(t) := u_0^\star$. 
Unfortunately, it is challenging to solve the Wasserstein DR-MPC problem due to the distributionally robust risk constraint~\eqref{DRMPCcons5}. 
This risk itself involves an optimization problem, which is infinite-dimensional.
To alleviate the computational difficulty, we reformulate the Wasserstein DR-MPC problem in a tractable form and propose efficient algorithms for solving the reformulated problem in the following sections.

\section{Finite-Dimensional Reformulation via Kantorovich Duality}\label{sec:reform}

To develop a computationally tractable approach to solving the Wasserstein DR-MPC problem, we propose a set of reformulation procedures. 
For ease of exposition, we suppress the subscripts in the DR-risk constraint \eqref{DRMPCcons5} and 
consider 
\begin{equation} \label{WDRconst}
\sup_{\mu \in \mathbb{D}} \cvar_\alpha^\mu [ \dist (y, \mathcal{Y} + w) ] \leq \delta.
\end{equation}

\subsection{Distance to the Safe Region}

The first step is to derive a simple expression for the loss of safety, $\dist (y,\mathcal{Y} + w)$.
Recall that the region occupied by an obstacle is represented as a convex polytope (via over-approximation if needed), i.e.,
\[
\mathcal{O}=\{ {y} \mid c_j^\top {y} \leq d_j, j=1, \ldots,m \}
\]
for some $c_j \in \mathbb{R}^{n_y}$ and $d_j \in \mathbb{R}$. 
Since $\mathcal{Y} = \mathbb{R}^{n_y} \setminus \mathcal{O}^o$,
the corresponding safe region can be expressed as the union of half spaces, i.e.,
\begin{equation}\label{safe}
\mathcal{Y}
:= \bigcup_{j=1}^m \{ {y} \mid c_j^\top {y} \geq d_j\}.
\end{equation}

From \eqref{safe} we see that the safe region is a union of halfspaces, resulting in the next lemma.

\begin{lemma}\label{lem:halfsp}
Suppose that the safe region is given by \eqref{safe}.
Then, the loss of safety \eqref{loss} can be expressed as
\[
\dist (y, \mathcal{Y}+w) = \bigg [\min_{j= 1, \ldots, m}
 \frac{d_j - c_j^\top (y-w)}{\|c_j \|_2}
\bigg ]^+.
\]
\end{lemma}

\begin{figure}[t]
\centering
	\includegraphics[width=2.5in]{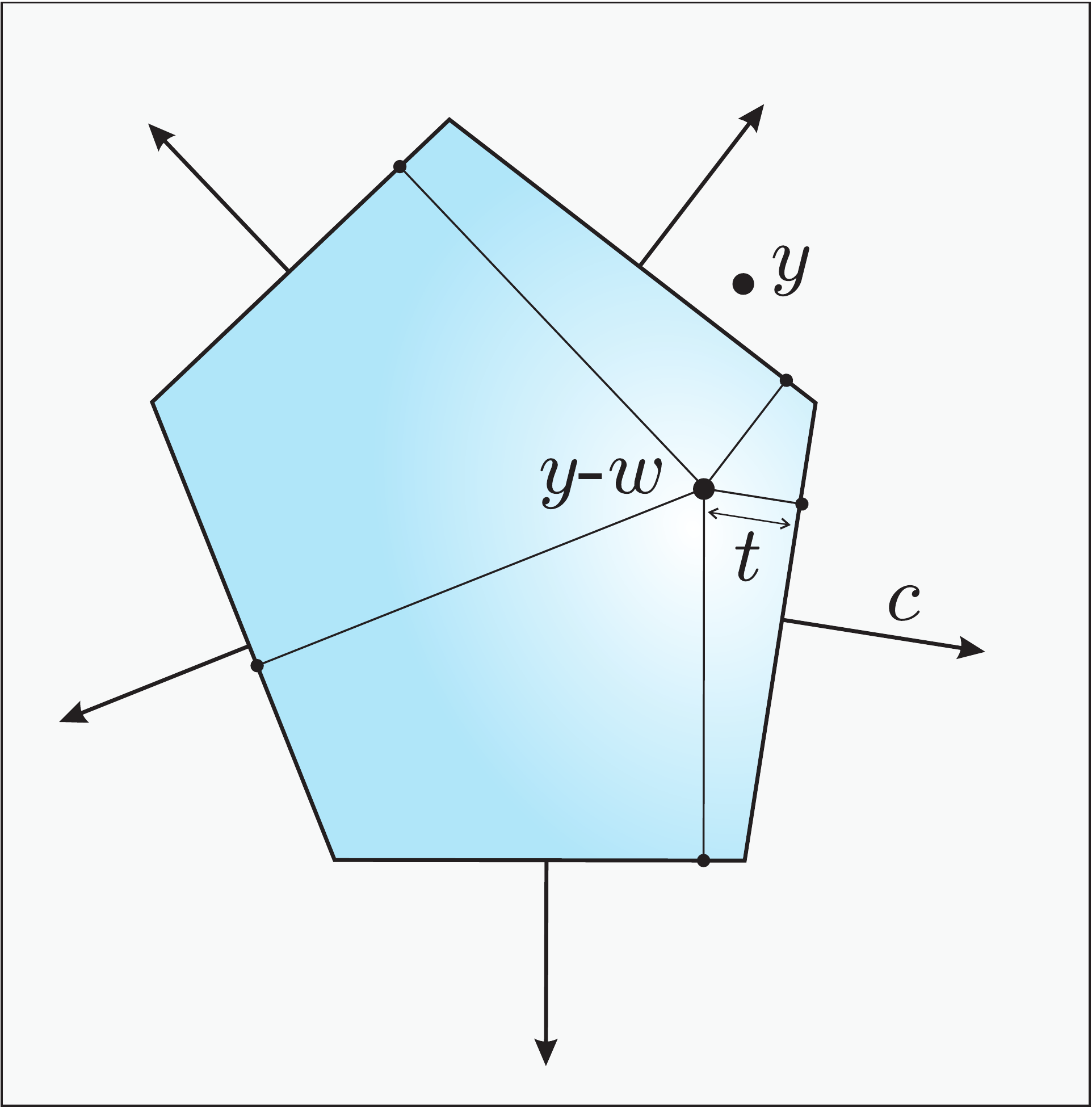}
	\caption{Illustration of the distance to the union of halfspaces.}
	\label{fig:Union}
\end{figure}

\begin{proof} 
First, we let 
\[
\mathcal{Y}_j :=\{y\mid c_j^\top y\geq d_j\}.
\]
Then, using the property that $\mathcal{Y}_j+w=\{y\mid c_j^\top (y-w)\geq d_j\}$, the distance between $y$ and each halfspace can be represented by
\begin{equation} \label{distj}
\mathrm{dist}(y,\mathcal{Y}_j+w) =\inf_{t}\{ \|t\|_2 \mid c_j^\top (y-t-w) \geq d_j\}.
\end{equation}
This equality is illustrated in Fig.~\ref{fig:Union}.
To derive the dual of
the optimization problem in \eqref{distj}, we first find the Lagrangian as
\[
L(t,\lambda)=\|t\|_2+\lambda \big[d_j-c_j^\top (y-t-w)\big].
\]
The corresponding dual function is obtained by
\begin{equation} \nonumber
\begin{split}
g(\lambda) &=\min_{t} \big \{\|t\|_2+\lambda\big[d_j-c_j^\top (y-t-w)\big] \big \}\\ 
&= \min_{t} \big \{\|t\|_2+\lambda c_j^\top t \big \}+\lambda\big[d_j-c_j^\top(y-w)\big].
\end{split}
\end{equation}
Note that 
\[
\min_t \{\|t\|_2 +\lambda c_j^\top t\} =\begin{cases} 0 \;& \mbox{if } \lambda \|c_j\|_2 \leq 1\\ -\infty \;& \mbox{otherwise.}
\end{cases}
\]
Therefore, the dual problem of \eqref{distj} can be derived as
\begin{align} \nonumber
\left\{\begin{array}{@{}l@{\;}l@{}} \max_{\lambda} & \lambda\big[d_j-c_j^\top(y-w)]\big]\\
\mbox{s.t.} & \lambda\|c_j\|_2\leq 1\\
\;&\lambda \geq 0
\end{array}\right.
\nonumber &=
\left\{\begin{array}{@{}l@{\;}l@{}}
\max_{\lambda} & \lambda\big[d_j-c_j^\top(y-w)\big]\\
\mbox{s.t.}& \lambda\leq \frac{1}{\|c_j\|_2}\\
\;&\lambda\geq 0
\end{array}\right.\\
&=\bigg[\frac{d_j-c_j^\top(y-w)}{\|c_j\|_2}\bigg]^+.\label{dual_dist}
\end{align}
The primal problem satisfies the refined Slater's conditions, as the inequality constraint is linear and the primal problem is feasible \cite[Section~5.2.3]{Boyd2004}. Therefore, we conclude that strong duality holds.

Now that we have the distance from a single halfspace, the distance from the safe region can be written as
\begin{align}
\mathrm{dist}(y,\mathcal{Y}+w)&=\min_{j=1, \ldots, m}\{\mathrm{dist}(y,\mathcal{Y}_j+w)\} \nonumber\\
&=\min_{j= 1, \ldots, m} \bigg \{
\frac{\big[d_j - c_j^\top (y-w)\big]^+}{\|c_j \|_2}
\bigg \},\nonumber
\end{align}
where the second equality follows directly from~\eqref{dual_dist}. 
This concludes the proof because minimum and $(\cdot)^+$ are interchangeable.
\end{proof}

\subsection{Reformulation of Distributionally Robust Risk Constraints}

The next step is to reformulate the distributionally robust risk constraint~\eqref{WDRconst} in a conservative manner. This reformulation will then be suitable for our purpose of limiting safety risk. 

\begin{lemma}\label{lem:cvar}
Suppose that the safe region is given by \eqref{safe}.
Then, the distributionally robust safety risk is upper-bounded as follows:
\begin{align} \nonumber
  &\sup_{\mu \in \mathbb{D}} \mathrm{CVaR}_\alpha^\mu [\dist (y, \mathcal{Y} + w) ] \leq  \inf_{z \in \mathbb{R}}z + \frac{1}{1-\alpha} \sup_{\mu\in\mathbb{D}} \mathbb{E}^\mu \Big[\max \big\{\min_{j}p_j(y,w)-z, -z,0 \big\} \Big ],\nonumber
\end{align}
where $p_j(y,w)=\frac{d_j-c_j^\top (y - w)}{\|c_j\|_2}$.
\end{lemma}

\begin{proof}
By the definition of CVaR and Lemma~\ref{lem:halfsp}, we have
\begin{align*}
\mathrm{CVaR}_\alpha^\mu [\dist (y, \mathcal{Y} + w) ]
   & = \inf_{z \in \mathbb{R}} \mathbb{E}^\mu \bigg [
z + \frac{\big (\dist (y,\mathcal{Y}+w) - z \big )^+}{1-\alpha}\bigg]\\
&=\inf_{z \in \mathbb{R}} \mathbb{E}^\mu\bigg [z + \frac{1}{1-\alpha}\Big (\big[\min_{j} p_j(y,w)\big]^+ - z\Big )^+
\bigg ].
\end{align*}
By the minimax inequality, we obtain that
\begin{align*}
  \sup_{\mu \in \mathbb{D}} \mathrm{CVaR}_\alpha^\mu [\dist (y, \mathcal{Y} + w) ]
    &\leq\inf_{z \in \mathbb{R}} \sup_{\mu\in\mathbb{D}} \mathbb{E}^\mu \bigg[ z + \frac{1}{1-\alpha}\Big (\big[\min_{j}p_j(y,w)\big]^+ - z\Big )^+\bigg ]\\
  &=\inf_{z \in \mathbb{R}} \sup_{\mu\in\mathbb{D}} \mathbb{E}^\mu \bigg[ z + \frac{1}{1-\alpha}\max \big \{ \min_j p_j (y, w) - z, -z, 0 \big \}
  \bigg ],
\end{align*}
and therefore the result follows.
\end{proof}
The upper-bound of the worst-case CVaR in Lemma~\ref{lem:cvar} is still difficult to evaluate because 
its inner maximization problem involves optimization over a set of distributions.
To resolve this issue, we use Wasserstein DRO based on Kantorovich duality to transform it into a finite-dimensional optimization problem as follows.

\begin{proposition}\label{prop:cvar}
Suppose that the uncertainty set is a compact convex polytope, i.e. $\mathbb{W} :=\{w\in \mathbb{R}^{n_y} \mid Hw\leq h\}$, where $H \in \mathbb{R}^{q \times n_y}$ and $h \in \mathbb{R}^q$. Then, the following equality holds:
\begin{equation}\nonumber
\begin{split}
 \sup_{\mu\in\mathbb{D}} \quad &\mathbb{E}^\mu \Big[ \max \Big \{\min_{j=1, \ldots, m}p_j(y,w)-z, -z,0 \Big \}\Big ] = \\
  \inf_{\lambda,s,\rho, \gamma, \eta,\zeta} \; & \lambda \theta+\sum_{i=1}^{N} s_i \\
  \mbox{\text{s.t.}} \; &\langle \rho_i,G(y - \hat{w}^{(i)})+g\rangle+\langle \gamma_i,h-H\hat{w}^{(i)}\rangle\leq s_i+z\\
  &\langle \eta_i,h-H\hat{w}^{(i)}\rangle\leq s_i+z\\
  &\langle \zeta_i,h-H\hat{w}^{(i)}\rangle\leq s_i\\
  &\|H^\top\gamma_i-G^\top\rho_i\|_*\leq\lambda\\
  &\|H^\top\eta_i\|_*\leq\lambda\\
  &\|H^\top\zeta_i\|_*\leq\lambda\\
  &\langle\rho_i,e_m\rangle=1\\
  &\gamma_{i}\geq 0,\; \rho_i\geq 0, \; \eta_{i}\geq 0,\; \zeta_{i}\geq 0,
\end{split}
\end{equation}
where all the constraints hold for $i=1,\dots,N$, and
 the dual norm $\|\cdot\|_*$ is defined by $\| z\|_*:= \sup_{\| \xi \| \leq 1} \langle z, \xi \rangle$. Here, $G \in \mathbb{R}^{m \times n_y}$ is a matrix with rows $-\frac{c_j^\top}{\|c_j\|_2}$, $j=1, \ldots, m$, $g \in \mathbb{R}^m$ is a column vector with entries $\frac{d_j}{\|c_j\|_2}$, $j = 1, \ldots, m$, and $e_m \in \mathbb{R}^m$ is a vector of all ones.
\end{proposition}

\begin{proof}
By the Kantorovich duality principle, we can rewrite the upper-bound of the worst-case CVaR in Lemma~\ref{lem:cvar}
in the following dual form:
\begin{equation} \nonumber
\begin{split}
  &\sup_{\mu\in\mathbb{D}} \mathbb{E}^\mu \Big [ \max \Big \{\min_{j=1,\dots,m}p_j(y,w)- z, -z,0 \Big \} \Big ] = \\
  &\inf_{\lambda \geq 0} \Big [ \lambda \theta + \frac{1}{N}\sum_{i=1}^N \sup_{w\in\mathbb{W}} \Big [ \max \Big \{\min_{j=1,\dots,m}p_j(y,w)- z, -z,0 \Big \}  - \lambda \| w - \hat{w}^{(i)} \| \Big ] \Big].
\end{split}
\end{equation}
It is proved in~\cite[Theorem~1]{Gao2016} that strong duality holds.
Introducing new auxiliary variable $s$ and following the procedure in \cite{Esfahani2018}, the dual problem above can be expressed as
\begin{align} \nonumber
&\left\{\begin{array}{@{}l@{\;}l@{}}
\inf_{\lambda, s}\; & \lambda\theta+\frac{1}{N}\sum_{i=1}^{N}s_i\\
\mbox{s.t.} \; & \sup_{w\in\mathbb{W}}\Big[ \max \big \{\min_{j} p_j(y,w)-z,-z,0 \big \} - \lambda \| w - \hat{w}^{(i)} \| \Big]\leq s_i\\
& \lambda \geq 0
\end{array}\right.\\
=&\left\{\begin{array}{@{}l@{\;}l@{}}
\inf_{\lambda, s}\; & \lambda\theta+\frac{1}{N}\sum_{i=1}^{N}s_i\\
\mbox{s.t.} \; & \sup_{w\in\mathbb{W}}[- \max_{\|\xi_{i,1}\|_*\leq \lambda} \langle \xi_{i,1},w - \hat{w}^{(i)}\rangle + \min_{j}p_j(y,w)]-z\leq s_i\\
& \sup_{w\in\mathbb{W}}[ - \max_{\|\xi_{i,2}\|_*\leq \lambda} \langle \xi_{i,2},w - \hat{w}^{(i)}\rangle]-z\leq s_i\\
& \sup_{w\in\mathbb{W}}[ - \max_{\|\xi_{i,3}\|_*\leq \lambda} \langle \xi_{i,3},w - \hat{w}^{(i)}\rangle ]\leq s_i\\
& \lambda \geq 0,
\end{array}\right.\nonumber
\end{align}
where the constraints hold for all $i$. 
In the second problem, we decompose the expression inside maximum and employ the definition of dual norm. Thereafter, since the set $\{\xi_{i,k}\mid\|\xi_{i,k}\|_*\leq\lambda\}$ is compact for any $\lambda\geq 0$, the minimax theorem can be used to rewrite the problem as
\begin{align} \nonumber
&\left\{\begin{array}{@{}l@{\;}l@{}}
\inf_{\lambda, s}\; & \lambda\theta+\frac{1}{N}\sum_{i=1}^{N}s_i\\
\mbox{s.t.} \; & \min_{\|\xi_{i,1}\|_*\leq \lambda}\sup_{w\in\mathbb{W}}[- \langle \xi_{i,1},w - \hat{w}^{(i)}\rangle + \min_{j} p_j(y,w)]-z\leq s_i\\
&\min_{\|\xi_{i,2}\|_*\leq \lambda} \sup_{w\in\mathbb{W}}[ -  \langle \xi_{i,2},w - \hat{w}^{(i)}\rangle]-z\leq s_i\\
&\min_{\|\xi_{i,3}\|_*\leq \lambda} \sup_{w\in\mathbb{W}}[ -  \langle \xi_{i,3},w - \hat{w}^{(i)}\rangle ]\leq s_i\\
& \lambda \geq 0
\end{array}\right.\\
=&\left\{\begin{array}{@{}l@{\;}l@{}}
\inf_{\lambda, s,\xi}\; & \lambda\theta+\frac{1}{N}\sum_{i=1}^{N}s_i\\
\mbox{s.t.} \; & \sup_{w\in\mathbb{W}}[\langle \xi_{i,1},w\rangle + \min_{j}p_j(y,w)] -\langle \xi_{i,1},\hat{w}^{(i)}\rangle-z\leq s_i\\
&\sup_{w\in\mathbb{W}} \langle \xi_{i,2},w \rangle-\langle \xi_{i,1},\hat{w}^{(i)}\rangle-z\leq s_i\\
&\sup_{w\in\mathbb{W}} \langle \xi_{i,3},w \rangle-\langle \xi_{i,1},\hat{w}^{(i)}\rangle\leq s_i\\
& \|\xi_{i,k}\|_*\leq\lambda,\;k=1,2,3,
\end{array}\right.\nonumber
\end{align}
where the constraints hold for all $i$. 
The first constraint can be written as sum of a conjugate function and the support function $\sigma_{\mathbb{W}}(\nu_i):= \sup_{w \in \mathbb{W}} \langle \nu_i, w \rangle$ since $-p_j(y,w)$ is proper, convex and lower semicontinuous. Likewise, the next two constraints can be represented using $\sigma_\mathbb{W}(\xi_{i,2})$ and $\sigma_\mathbb{W}(\xi_{i,3})$ as follows:
  \begin{align}
   &
\left\{\begin{array}{@{}l@{\;}l@{}} 
   \inf_{\lambda,s,\xi,\nu} & \lambda \theta + \sum_{i=1}^N s_i\\
   \mbox{$\textrm{s.t.}$}& \sup_{w} [ \langle \xi_{i,1}-\nu_{i},w\rangle + \min_{j}p_j(y,w) ] +\sigma_{\mathbb{W}}(\nu_{i})-\langle \xi_{i,1}, \hat{w}^{(i)}\rangle -z\leq s_i\\
   & \sigma_{\mathbb{W}}(\xi_{i,2})-\langle \xi_{i,2}, \hat{w}^{(i)}\rangle-z \leq s_i\\
   & \sigma_{\mathbb{W}}(\xi_{i,3})-\langle \xi_{i,3}, \hat{w}^{(i)}\rangle \leq s_i\\
   & \|\xi_{i,k}\|_*\leq\lambda, \; k=1,2,3,
\end{array}\right.\label{epi_const}
\end{align}
where the constraints hold for all $i$.

On the other hand, we note that
\begin{equation} \nonumber
\begin{split}
\sup_{w} \Big [ \langle \xi_{i,1}-\nu_{i},w\rangle + \min_{j =1,\ldots, m}p_j(y,w) \Big ]
&=\left\{\begin{array}{@{}ll}
\sup_{w,\tau} \;& \langle \xi_{i,1}-\nu_{i},w\rangle+\tau\\
\mbox{s.t.} \;& G (y - w)+g\geq \tau e 
\end{array}\right.\\
&=\left\{\begin{array}{@{}ll}
\inf_{\rho_i} \; & \langle \rho_i,g+Gy\rangle\\
\mbox{s.t.} \; & G^\top\rho_i= \xi_{i,1} - \nu_{i}\\
& \langle \rho_i,e_m\rangle=1\\
&\rho_i\geq 0,
\end{array}\right.
\end{split}
\end{equation}
where the last equality follows from strong duality of linear programming, which holds because the primal maximization problem is feasible.
By the definition of support functions, we also have
\[
\sigma_{\mathbb{W}}(\nu_i)=\left\{\begin{array}{@{}ll} \sup_{w} \; &\langle \nu_i,w \rangle\\ \mbox{s.t.}\; & Hw\leq h\end{array}\right.=\left\{\begin{array}{@{}ll}\inf_{\gamma_i} \; &\langle \gamma_i, h \rangle\\ \mbox{s.t.}\; & H^\top \gamma_i = \nu_i\\
& \gamma_i\geq 0,\end{array}\right.
\]
where the last equality follows from strong duality of linear programming, which holds since the uncertainty set is nonempty. Similar expressions are derived for $\sigma_\mathbb{W}(\xi_{i,2})$ and $\sigma_\mathbb{W}(\xi_{i,3})$ with Lagrangian multipliers $\eta_i$ and $\zeta_i$, respectively.
By substituting the results above into \eqref{epi_const}, we conclude that the proposed reformulation is exact.
\end{proof}

\subsection{Reformulation of the Wasserstein DR-MPC Problem}

We are now ready to reformulate the Wasserstein DR-MPC problem~\eqref{DRMPC} as a finite-dimensional optimization problem
by using Lemma~\ref{lem:cvar} and Proposition~\ref{prop:cvar}.
Putting all the pieces in Lemma~\ref{lem:cvar} and Proposition~\ref{prop:cvar} together into \eqref{DRMPC}, we have
\begin{subequations}\label{WDR-MPC}
\begin{align}
\inf_{\substack{\bold{u}, \bold{x},\bold{y},\bold{z},\lambda,\\\bold{s},\bold{\rho},\bold{\gamma},\bold{\eta},\bold{\zeta}}} \; & J(x(t), \bold{u}) := \sum_{k=0}^{K-1} r (x_k, u_k)+q(x_K) \\
\mbox{s.t.} \; & x_{k+1} = f (x_k,u_k) \\
& y_k=h(x_k,u_k)\\
& x_0 = x(t) \\
&z_{\ell,k}+\frac{1}{1-\alpha}\Big[\lambda_{\ell,k} \theta+\frac{1}{N_k}\sum_{i=1}^{N_k} s_{\ell,k,i}\Big]\leq \delta_\ell \\
&\langle\rho_{\ell,k,i},G_t(y_k - \hat{w}^{(i)}_{\ell,t,k})+g_t\rangle  +\langle \gamma_{\ell,k,i},h-H\hat{w}_{\ell,t,k}^{(i)}\rangle\leq s_{\ell,k,i}+z_{\ell,k} \label{bl_const}\\
&\langle \eta_{\ell,k,i},h-H\hat{w}_{\ell,t,k}^{(i)}\rangle\leq s_{\ell,k,i}+z_{\ell,k} \\
&\langle \zeta_{\ell,k,i},h-H\hat{w}_{\ell,t,k}^{(i)}\rangle\leq s_{\ell,k,i} \\
&\|H^\top\gamma_{\ell,k,i}-G_t^\top\rho_{\ell,k,i}\|_*\leq\lambda_{\ell,k} \\
&\|H^\top\eta_{\ell,k,i}\|_*\leq\lambda_{\ell,k} \\
&\|H^\top\zeta_{\ell,k,i}\|_*\leq\lambda_{\ell,k}\\
&\langle\rho_{\ell,k,i},e_m\rangle=1 \\
&\gamma_{\ell,k,i},\rho_{\ell,k,i}, \eta_{\ell,k,i},\zeta_{\ell,k,i}\geq 0 \\
&x_k\in\mathcal{X},\; u_k\in\mathcal{U}, \; z_{\ell,k}\in\mathbb{R}, 
\end{align}
\end{subequations}
where all the constraints hold for $k=1,\dots,K$, $\ell = 1, \ldots, L$ and $i=1,\dots,N_k$, except for the first constraint and $u_k \in \mathcal{U}$, which should hold for $k=0,\dots,K-1$, and the second constraint and $x_k\in\mathcal{X}$, which should hold for $k=0,\dots,K$.

The overall motion control process is as follows. First, at stage $t$ the initial state $x_0$ in MPC is set to be the current state $x(t)$. Also, the current safe region $\mathcal{Y}_\ell(t)$ is observed to return $G_t$ and $g_t$. Second, the Wasserstein DR-MPC problem~\eqref{WDR-MPC} is solved to find a solution $\bold{u}^\star$ satisfying the risk constraint even when the actual distribution deviates from the empirical distribution~\eqref{emp} within the Wasserstein ball~\eqref{ball}. 
Then, the first component of the optimal control input sequence $u_0^\star$ is selected as the control input at stage $t$ and applied to the robotic vehicle. 
These two steps are repeated for all time stages until the desired position in the configuration space is reached.

The proposed reformulation resolves the infinite-dimensionality issue in the original Wasserstein DR-MPC problem. 
Thus, the reformulated problem is easier to solve than the original one. 
However, it is still
nonconvex due to the nonlinear system dynamics and output equations, as well as the bilinearity of the fifth constraint~\eqref{bl_const}; all the other constraints and the objective function are convex. 
Thus, a locally optimal solution can be found by using efficient nonlinear programming (NLP) algorithms such as interior-point methods, sequential quadratic programming, etc~\cite{Nocedal2006}. 
However, in some specific cases, e.g., when the system dynamics and the output equations are affine, we can use relaxation techniques to find a globally optimal solution. 
One  such relaxation method will be discussed in the following section.

\section{Spatial Branch-and-Bound Algorithm Using McCormick Envelopes}\label{sec:sBB}

In general, solving the nonconvex problem \eqref{WDR-MPC} is a nontrivial task. 
As mentioned, 
if a general NLP algorithm is used to directly solve the problem, then the solution returned by it may not be optimal due to multiple possible local minima. 
However, for the case of affine system dynamics and output equations, we develop an efficient approach to obtaining a globally optimal solution to the DR-MPC problem. 
Following the techniques introduced in~\cite{liberti2008introduction}, we use McCormick envelopes~\cite{mccormick1976computability} to relax the nonconvex bilinear constraint~\eqref{bl_const} and form an under-estimate of the original DR-MPC problem. 
Then, an $\varepsilon$-global optimum is found by using the sBB algorithm.

In order to relax the bilinear constraint~\eqref{bl_const}, we first introduce a new variable $\chi_{\ell,k,i}\in\mathbb{R}^{m\times n_y}$. For notational simplicity, we omit the subscript $t$ on $G_t$ and the subscripts $\ell,k,i$ on the other variables. 
The following new equality constraint is added to the optimization problem:
\begin{equation}
\chi_{(p,q)}=\rho_{(p)} G_{(p, q)} y_{(q)},\label{eq_bef}
\end{equation}
where $\rho_{(p)}$ is the $p$th element of $\rho$, $y_{(q)}$ is the $q$th element of $y$, and $\chi_{(p,q)}$ and $G_{(p,q)}$ are the elements on the $p$th row and the $q$th column of $\chi$ and $G$, respectively.
Then, the bilinear constraint \eqref{bl_const} can be expressed using \eqref{eq_bef} as follows:
\[
e_m^\top \chi e_{n_y} -\rho^\top G\hat{w}+\rho^\top g+\gamma^\top(h-H\hat{w})\leq s+z,
\]
where $e_n$ denotes the $n$-dimensional column vector of all ones.

It is clear that the new inequality constraint is convex. However, we still have the equality constraint~\eqref{eq_bef}, which is bilinear. 
We therefore use the McCormick envelope to find a convex relaxation of the bilinear constraint.

\begin{proposition}[McCormick Relaxation]
Suppose that
\begin{equation}\nonumber
\begin{split}
&\underline{\rho} \leq\rho \leq \overline{\rho}\\
&\underline{y} \leq y\leq \overline{y}
\end{split}
\end{equation}
 for some $\underline{\rho}, \overline{\rho} \in \mathbb{R}^m$ and $\underline{y}, \overline{y} \in \mathbb{R}^{n_y}$.
 Then, the equality constraint \eqref{eq_bef} can be relaxed as the following set of inequality constraints:
\begin{subequations}
\begin{align}
&\chi_{(p,q)}\geq \underline{\rho}_{(p)} G_{(p,q)} y_{(q)}+\rho_{(p)} G_{(p,q)} \underline{y}_{(q)}- \underline{\rho}_{(p)} G_{(p,q)} \underline{y}_{(q)} \label{under1}\\
&\chi_{(p,q)}\geq \overline{\rho}_{(p)} G_{(p,q)} y_{(q)}+\rho_{(p)} G_{(p,q)} \overline{y}_{(q)}- \overline{\rho}_{(p)} G_{(p,q)} \overline{y}_{(q)} \label{under2}\\
&\chi_{(p,q)}\leq \overline{\rho}_{(p)} G_{(p,q)} y_{(q)}+\rho_{(p)}G_{(p,q)} \underline{y}_{(q)} -\overline{\rho}_{(p)} G_{(p,q)} \underline{y}_{(q)} \label{over1}\\
&\chi_{(p,q)}\leq \rho_{(p)} G_{(p,q)} \overline{y}_{(q)}+ \underline{\rho}_{(p)} G_{(p,q)} y_{(q)}- \underline{\rho}_{(p)} G_{(p,q)} \overline{y}_{(q)} \label{over2}\\
&\underline{\rho}_{(p)} \leq\rho_{(p)}\leq \overline{\rho}_{(p)}\label{bound1}\\
&\underline{y}_{(q)} \leq y_{(q)}\leq \overline{y}_{(q)},\label{bound2}
\end{align}\label{McCormick}
\end{subequations}
which hold for all $(p, q) \in \mathbb{R}^m \times \mathbb{R}^{n_y}$.
\end{proposition}

\begin{figure}[t]
\begin{subfigure}{0.5\linewidth}
  \centering
  \includegraphics[width=3in]{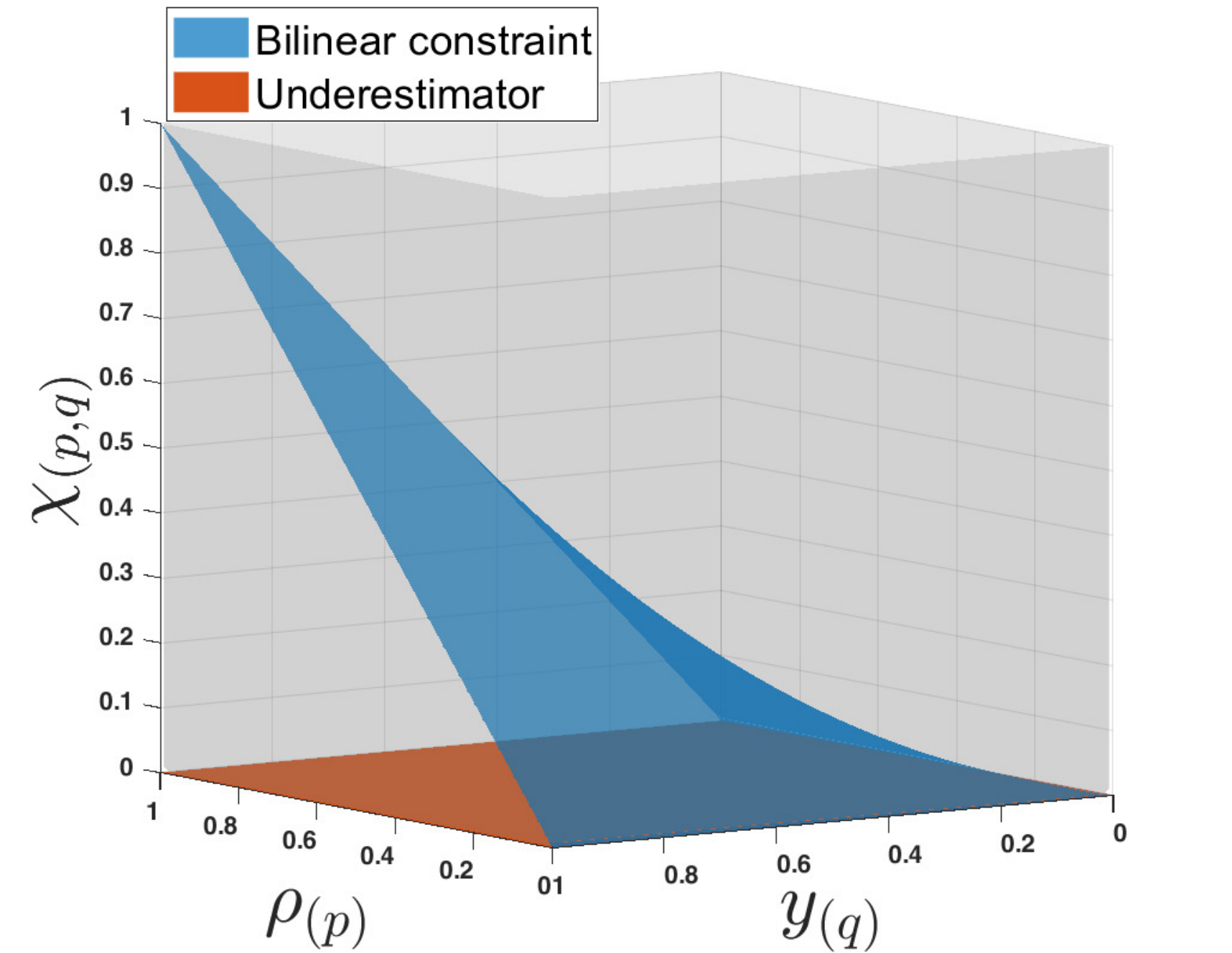}
  \caption{ }
  \label{fig:McCormicka}
\end{subfigure}%
\begin{subfigure}{0.5\linewidth}
  \centering
  \includegraphics[width=3in]{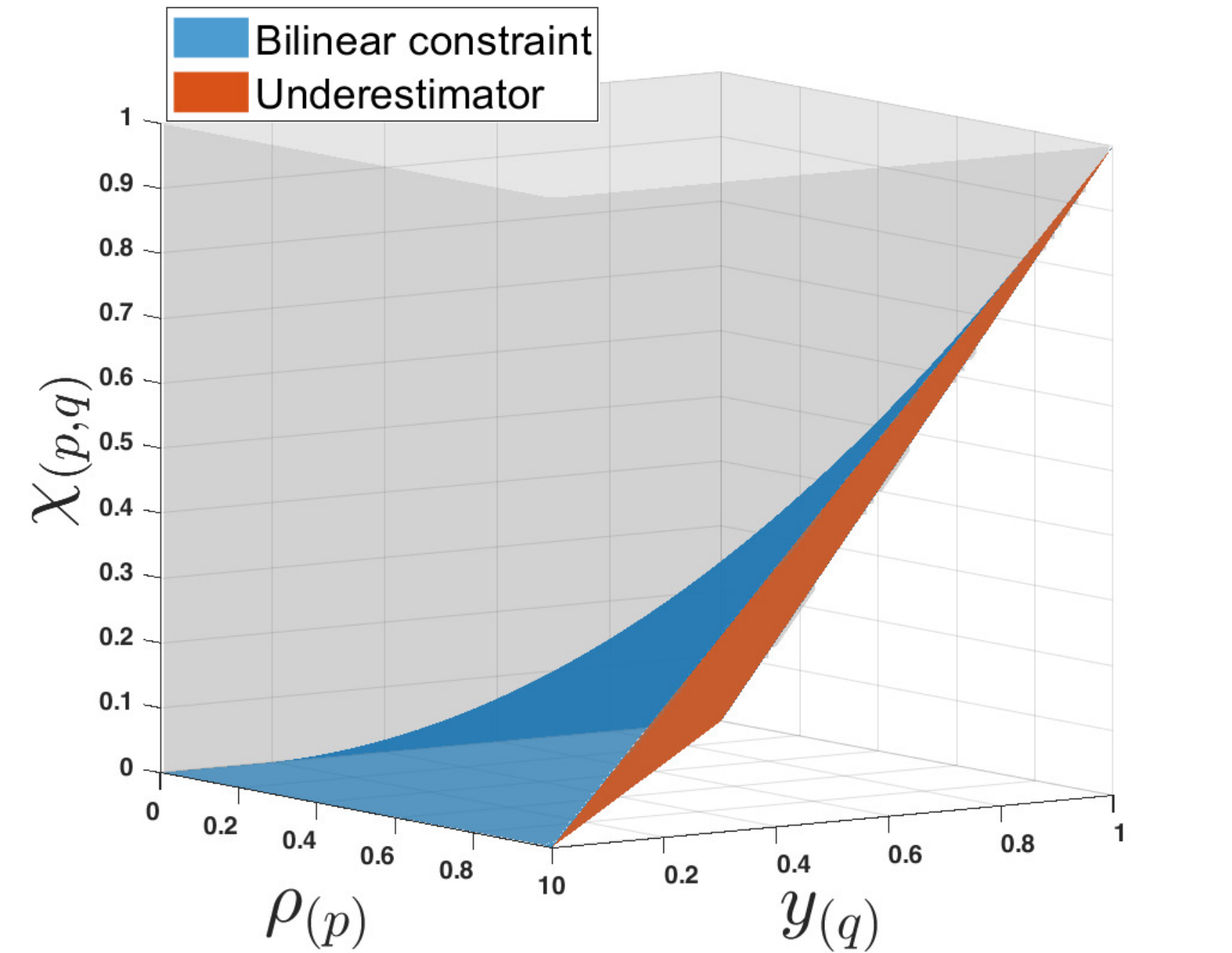}
  \caption{ }
  \label{fig:McCormickb}
\end{subfigure}
\begin{subfigure}{0.5\linewidth}
  \centering
  \includegraphics[width=3in]{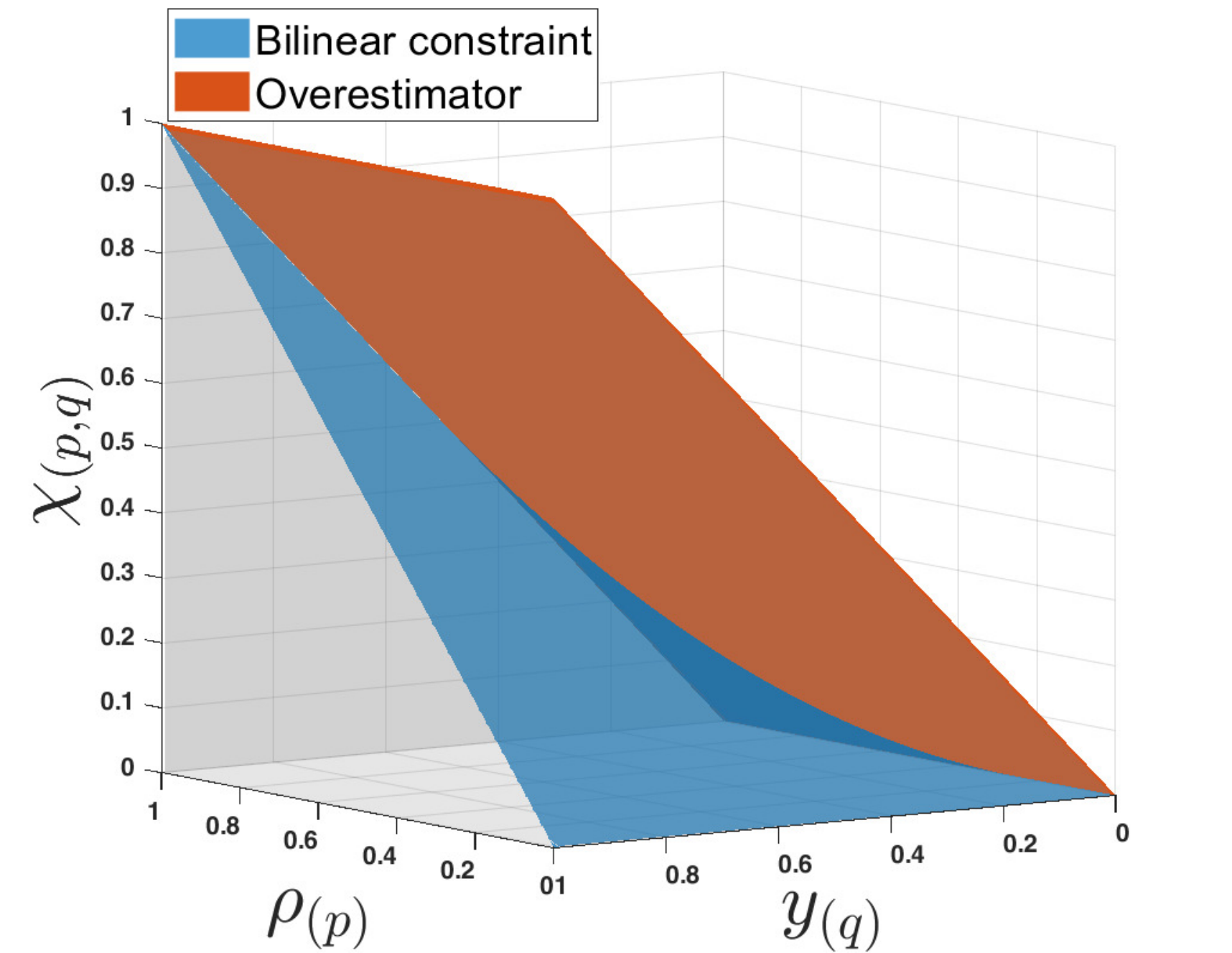}
  \caption{ }
  \label{fig:McCormickc}
\end{subfigure}%
\begin{subfigure}{0.5\linewidth}
  \centering
  \includegraphics[width=3in]{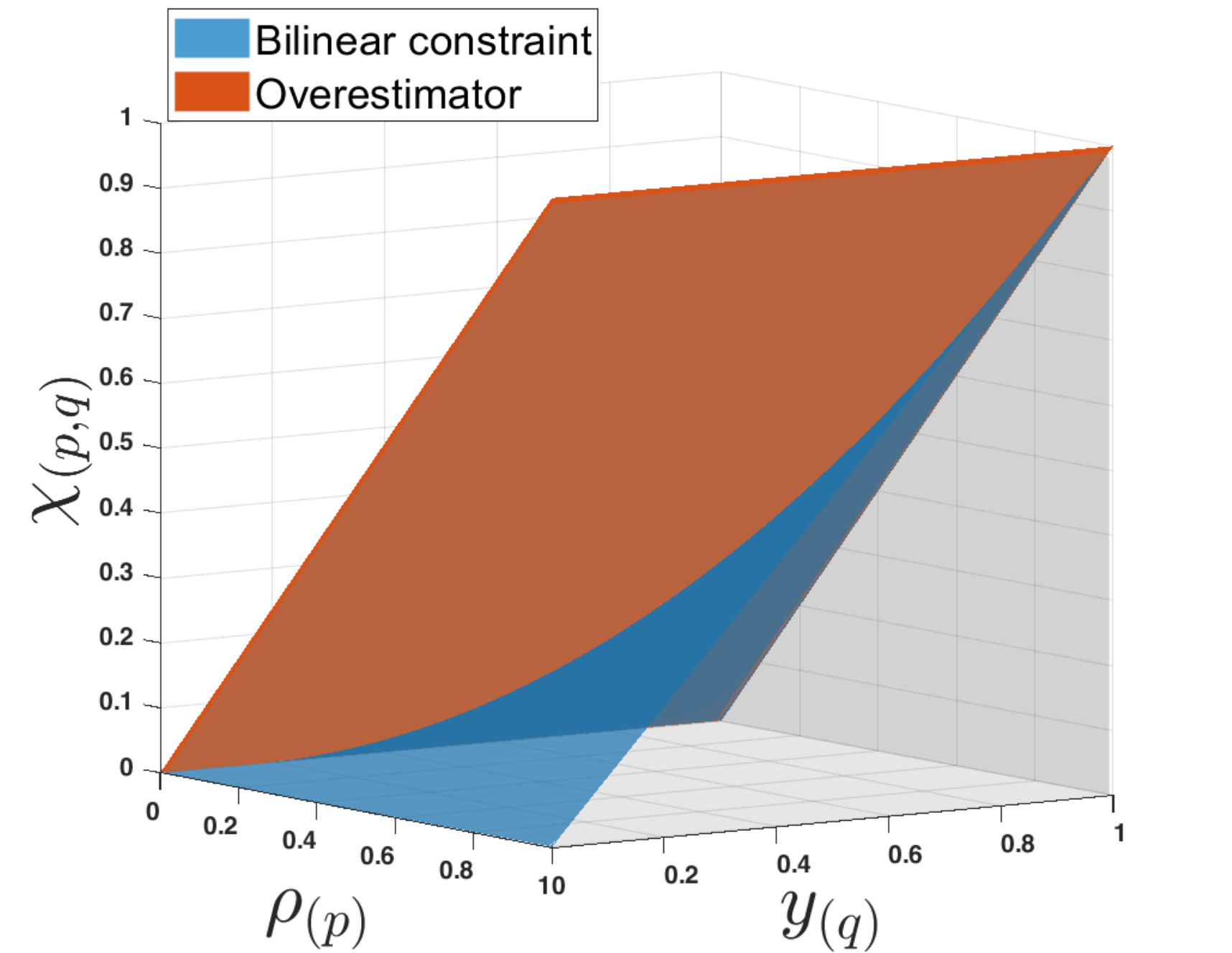}
  \caption{ }
  \label{fig:McCormickd}
\end{subfigure}
\caption{McCormick relaxation of the bilinear constraint~\eqref{bl_const}. (a) and (b) show the bilinear constraint (in blue), its underestimators (in orange) and the feasible regions (in gray) corresponding to the inequality constraints \eqref{under1} and \eqref{under2}, respectively. The overestimators (in orange) and the corresponding feasible regions (in gray) for \eqref{over1} and \eqref{over2} are plotted in (c) and (d), respectively. (b) and (d) are rotated $180^\circ$ around the $z$-axis for better visibility. }
\label{fig:McCormick}
\end{figure}

An example of the McCormick envelope on $\chi_{(p,q)}$ assuming $(\underline{\rho}_{(p)}, \overline{\rho}_{(p)}, \underline{y}_{(q)}, \overline{y}_{(q)})= (0,1, 0,1)$
 is shown in Fig~\ref{fig:McCormick}. Fig. \ref{fig:McCormicka} and \ref{fig:McCormickb} demonstrate the underestimators of $\chi_{(p,q)}$ defined in \eqref{under1} and \eqref{under2}, respectively, while Fig. \ref{fig:McCormickc} and \ref{fig:McCormickd} show the overestimators of $\chi_{(p,q)}$ defined in \eqref{over1} and \eqref{over2}, respectively. Each shaded area in the figures represents the region of the variables satisfying the corresponding inequality constraint taking into account the constraints \eqref{bound1} and \eqref{bound2}. Altogether these constraints form the convex hull of the original feasible set. 
 
After such relaxation, the nonconvex optimization problem can be solved using the sBB algorithm, where the range of each variable is divided into multiple subregions. For each subregion, we aim to find the global optimum by evaluating the upper and lower bounds of the objective function value. The upper bound is found by solving the original nonconvex problem, while for the lower bound we solve a convex optimization problem based on the McCormick relaxation. The global optimum of the subregion is attained when the bounds converge, i.e. the difference between the upper and lower bounds meets the required tolerance level. Otherwise, the region is divided into smaller subregions and the steps are repeated. This method is based on the idea of ``divide and conquer," where each of the subproblems is smaller than the parent problem. 

\begin{algorithm}[t]
\SetKw{Input}{Input:}
\Input $\varepsilon>0, \underline{\varphi}, \overline{\varphi}$;\\
$\overline{v}\leftarrow \infty$\;
$\varphi^\star\leftarrow [\infty,\dots,\infty]$\;
Apply optimization-based bound tightening\;
$\mathcal{L} \leftarrow \{ [\underline{\varphi}, \overline{\varphi}] \}$\;
\While{$\mathcal{L}\neq \emptyset$}{
Choose $\Phi \in \mathcal{L}$\;
$\mathcal{L} \leftarrow \mathcal{L} \setminus \{\Phi\}$\;
Solve the relaxed problem on $\Phi$,\\ and select $(\underline{v}, \varphi^r)$ as an optimal value-solution pair\;
$\underline{v}_\Phi \leftarrow \underline{v}$\;
\If{$\underline{v} \leq \overline{v}$ and $\underline{v}\neq \infty$}{
Solve original problem on $\Phi$,\\
and select $(\varphi^{new}, \overline{v}^{new})$ as a locally optimal value-solution pair;\\
\If{fail to find a solution}{
$\overline{v}^{new} \leftarrow \infty$\;
$\varphi^{new} \leftarrow [\infty,\dots,\infty]$\;
}
\If{$\overline{v}^{new} < \overline{v}$}{
$\varphi^\star \leftarrow \varphi^{new}$\;
$\overline{v} \leftarrow \overline{v}^{new}$\;
\For{\emph{all} $S \in \mathcal{L}$ with $\underline{v}_S > \overline{v}$}{
$\mathcal{L}\leftarrow \mathcal{L}\setminus \{S\}$;
}
\eIf{$\overline{v} - \underline{v} \leq \varepsilon$ }{
$\overline{v}$ is the global minimum of the region; \\
}
{
Branch $\Phi$ to $\Phi_{1}$ and $\Phi_{2}$\;
$\mathcal{L}\leftarrow \mathcal{L} \cup \{\Phi_1,\Phi_2\}$\;
$\underline{v}_{\Phi_1}\leftarrow \underline{v},\; \underline{v}_{\Phi_2}\leftarrow \underline{v}$\;
}
}
}
\Return $(\varphi^\star, \overline{v})$\;
}\caption{The sBB algorithm for the Wasserstein DR-MPC problem~\eqref{WDR-MPC}}\label{sBB}
\end{algorithm}

The overall sBB algorithm is shown in Algorithm~\ref{sBB}, where we let
$\varphi$ be the vector consisting of all variables, i.e.
\[
\varphi := [\bold{u},\bold{x},\bold{y},\bold{z},\bold{\lambda},\bold{s},\bold{\rho},\bold{\gamma},\bold{\eta},\bold{\zeta},\bold{\chi}].
\]
We also set initial values for the lower bound $\underline{\varphi}$ and the upper bound $\overline{\varphi}$ of $\varphi$.
As shown in Lines 2 and 3,
 the current best upper bound $\overline{v}$ of the objective function and the corresponding optimal solutions $\varphi^\star$ are initialized to infinity as we do not know these values yet. 
 Optionally, as suggested in~\cite{liberti2008introduction}, an optimization-based bound tightening can be performed to tighten $\underline{\varphi}$ and $\overline{\varphi}$, as initially they can be too loose (Line 4). 
 The list of regions $\mathcal{L}$ is initialized as a singleton, where the single region is
 the Cartesian product of all variable ranges, i.e., $\mathcal{L} \leftarrow \{[\underline{\varphi}, \overline{\varphi}] \}$ (Line~5).

\begin{figure*}[t]
\centering
\includegraphics[width=6.3in]{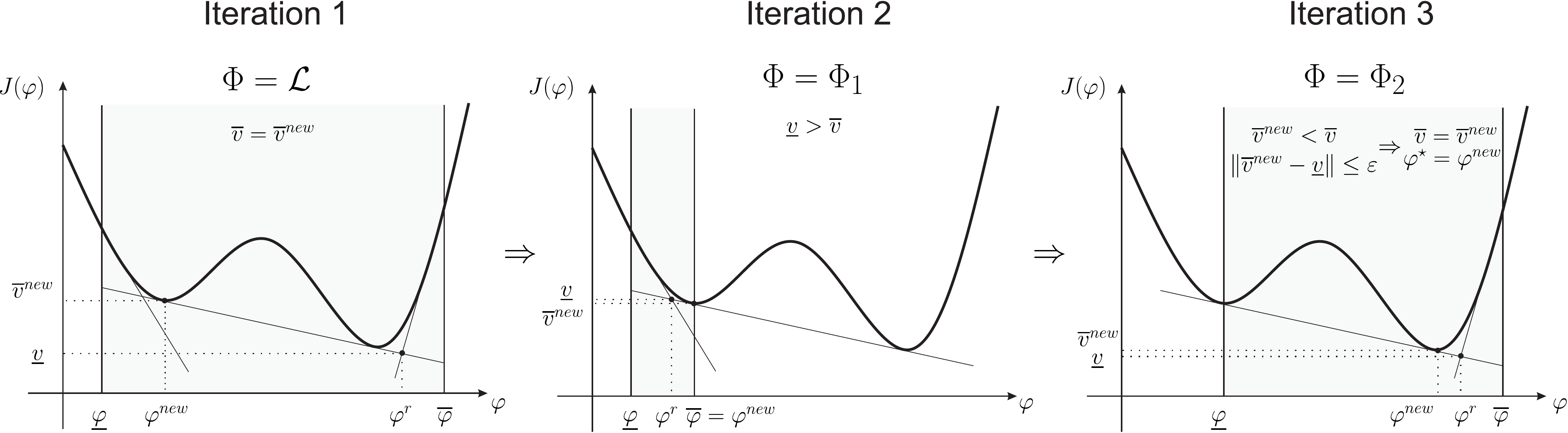}
\caption{Example of the sBB procedure for a simple one-dimensional nonconvex function $J(\varphi)$.}
\label{fig:sBB}	
\end{figure*}

In the main loop, each region in $\mathcal{L}$ is examined by first choosing a region $\Phi$ from the list and removing it (Lines 7-8).
Constraining all the variables to lie in $\Phi$, the convex program based on the McCormick relaxation is then solved. The optimal value of this convex program yields a lower bound (denoted by $\underline{v}$) to the objective function value of the original problem in that region with corresponding solution ${\varphi}^r$ (Lines 9-10).

If the relaxed problem is feasible and the lower bound $\underline{v}$ is smaller than the current best upper bound $\overline{v}$, then we proceed to the next step (Line 12). This means that we are in the correct direction, as no lower bound can be greater than the best upper bound. Otherwise, we go back to  Line 7 to select another region from the list $\mathcal{L}$, as the current region does not have any possibility of containing the global optimum.

In the next step, we attempt to solve the original problem constrained to be in the subregion $\Phi$. The objective function value $\overline{v}^{new}$ corresponding to the locally optimal solution $\varphi^{new}$ of the original problem is chosen as the upper bound to the original problem in $\Phi$ (Lines 13-14).

In general, when using an NLP solver, it might fail to find a locally optimal solution even though one might exist. 
In that case, the objective function value $\overline{v}^{new}$ and the corresponding solution $\varphi^{new}$ are set to infinity (Lines 15-17).

Now, having the new upper bound $\overline{v}^{new}$, we check whether it is better than the current best bound $\overline{v}$ (Line 18). If that is the case, $\overline{v}$ and $\varphi^\star$ are updated as the new upper bound $\overline{v}^{new}$ and the corresponding solution $\varphi^{new}$ (Lines 19-20). Here, we also prune (fathom) all the subregions in $\mathcal{L}$, which have a lower bound greater than $\overline{v}$ as they cannot contain the global optimum (Lines 21-22).

The global optimum for the region is attained if the difference between upper and lower bounds is smaller than a user-defined tolerance $\varepsilon>0$ (Line 23). In that case, the upper bound $\overline{v}$ is accepted as the global optimal value for the region $\Phi$ with corresponding optimal solution $\varphi^{new}$ (Line 24). This means that there is no need to further examine the region. Therefore, we go back to Line 7 to explore another region.

If the global optimum for the region is not found, the region is branched into smaller subregions $\Phi_1$ and $\Phi_2$, which are appended to the list $\mathcal{L}$ to further decrease the difference between the lower and upper bounds (Lines 26-27). The branching step is important for improving the convergence of the algorithm. One  widely used approach is to select  the branching variable by evaluating the original bilinear term \eqref{eq_bef} for $\rho^{new}$ and $y^{new}$ and computing its error from the variable $\chi^r$ in the relaxed problem. Then, the branching variable becomes the one resulting in the largest error.
Suppose that variable $x$ is selected as the branching variable. Then, the range of it is partitioned into $[\underline{x},x^{new}]$ and $[x^{new},\overline{x}]$. As a result, two new subregions $\Phi_1$ and $\Phi_2$ are obtained, and a lower bound of $\underline{v}$ is assigned to each of them (Line 28).

The above steps are repeated until the list of regions $\mathcal{L}$ becomes empty, meaning that all the subregions have been explored. As a result, the algorithm returns the optimal value $\overline{v}$ and the optimal solution $\varphi^\star$ of the original problem (Line 29).

An example of the sBB algorithm is illustrated in Fig.~\ref{fig:sBB}, where we want to minimize a nonconvex objective function $J(\varphi)$ over the region $[\underline{\varphi}, \overline{\varphi}]$. 
To begin with, $J(\varphi)$ is approximated by its secants, after which we proceed to the algorithm. 
In the first iteration, a lower bound $\underline{v}$ and an upper bound $\overline{v}^{new}$ of the optimal value are determined. 
Then, the current best upper bound is set to $\overline{v}$. The region $\Phi$ is divided into two subregions $\Phi_1=[\underline{\varphi}, \varphi^{new}]$ and $\Phi_2=[\varphi^{new}, \overline{\varphi}]$ afterwards. 
In the second iteration, the original and the relaxed problems are solved on $\Phi_1$. However, since $\underline{v} > \overline{v}$, the iteration terminates. 
In the third iteration, the second subregion $\Phi_2$ is considered. 
The new lower bound $\underline{v}$ and upper bound $\overline{v} = \overline{v}^{new}$ are better than the current best bounds. 
Thus, we check whether the lower and upper bounds are close to each other. 
Since the gap is less than the predefined tolerance $\varepsilon$, we conclude that $\overline{v} = \overline{v}^{new}$ is the global optimum of the function $J(\varphi)$ with optimal solution $\varphi^\star = \varphi^{new}$.

Note that the sBB algorithm finds an $\varepsilon$-optimal solution, unlike general NLP methods that only compute locally optimal solutions.
However, finding a locally optimal solution via an NLP algorithm is likely to be faster than obtaining a globally optimal solution via Algorithm \ref{sBB} due to the complexity of sBB. 
The fathoming in line 19 of Algorithm \ref{sBB} is a useful tool for accelerating the algorithm. Also, a good choice of subregion $\Phi$ in every iteration can improve the convergence. One of the widely accepted approaches is choosing the region with the ``best bound first'' rule, meaning that the subregion with the lowest lower bound is selected first~\cite{smith1997optimal}.

\section{Out-of-Sample Performance Guarantee}\label{sec:outofsample}

A notable advantage of the Wasserstein DR-MPC method is to assure a \emph{probabilistic out-of-sample performance guarantee}, meaning that the safety risk constraint is satisfied with probability no less than a certain threshold, even when evaluated under a set of new samples chosen independently of the training data.
This is a finite-sample (non-asymptotic) guarantee, which cannot be attained in many popular methods such as SAA.

Let $(\bold{u}^\star, \bold{x}^\star, \bold{y}^\star)$ denote an optimal solution to the Wasserstein DR-MPC problem~\eqref{DRMPC} at stage $t$, obtained by using the training dataset $\{\hat{w}_{\ell,t,k}^{(1)},\dots,\hat{w}_{\ell,t,k}^{(N_k)}\}$. Then, the \emph{out-of-sample risk} at stage $t$ is defined by
\begin{equation}
\cvar_\alpha^{\mu} [ \dist (y^\star(t+1), \mathcal{Y}(t) + w_{\ell,t,1}) ],\label{multistage_osp}
\end{equation}
which represents the risk of unsafety evaluated under the (unknown) true loss distribution $\mu$.
However, as $\mu$ is unknown in practice, it is impossible to exactly evaluate the out-of-sample risk. 
Instead, we seek a motion control solution that provides the following probabilistic performance guarantee:
\begin{equation}\label{Rel}
\begin{split}
\mu^{N_1} \Big \{
\cvar_\alpha^{\mu} [ \dist (y^\star(t+1), \mathcal{Y}(t) + w_{\ell,t,1}) ]\leq \delta_\ell
\Big \} \geq 1-\beta \quad \forall t, 
\end{split}
\end{equation}
where $\beta\in (0,1)$. 
This inequality represents that 
the risk of unsafety is no greater than the risk-tolerance parameter $\delta$ with $(1-\beta)$ confidence level. 
We refer to the probability on the left-hand side of~\eqref{Rel} as the \emph{reliability} of the motion control.
The reliability increases with the Wasserstein ball radius $\theta$. 
Thus, $\theta$ needs to be carefully determined to establish the probabilistic out-of-sample performance guarantee with desired $\beta$.

The required radius can be found from the following measure concentration inequality for Wasserstein ambiguity sets~\cite[Theorem 2]{Fournier2015}:\footnote{The measure concentration inequality assumes that $\mathbb{E}^\mu[ \exp ( \| w\|^c ) ] \leq B$ for $c > 1$ and $B > 0$, i.e. light-tailed distribution. In our problem formulation, this condition holds trivially for any compact uncertainty set $\mathbb{W}$.}
\begin{equation} \label{measure}
\begin{split}
&\mu^{N_1} \big \{\hat{w} \mid 
W (\mu, \nu) \geq \theta
\big \} \leq c_1 \big [ b_1(N_1, \theta) \bold{1}_{\{\theta\leq 1\}} + b_2(N_1, \theta) \bold{1}_{\{\theta > 1\}} \big ],
\end{split}
\end{equation}
where
\begin{equation} \nonumber
\begin{split}
b_1 (N, \theta) &:= \left \{
\begin{array}{ll}
\exp (-c_2 N \theta^2) & \mbox{if } n_y < 2\\
\exp (-c_2 N (\frac{\theta}{\log(2+1/\theta)})^2 ) & \mbox{if } n_y = 2\\
\exp (-c_2 N \theta^{n_y}  ) &\mbox{otherwise}
\end{array}
\right.\\
b_2 (N, \theta) &:= 
\exp ( -c_2 N \theta^{c} )
\end{split}
\end{equation}
for some constants $c_1, c_2 > 0$.
Suppose that the radius is chosen as
\[
\theta := \begin{cases} \Big[ \frac{\log (c_1 / \beta)}{c_2 N_1} \Big]^{1/c} & \text{if}\; N_1 < \frac{1}{c_2}\log (c_1/\beta)\\ \Big[ \frac{\log(c_1 / \beta)}{c_2 N_1} \Big]^{1/n_y} & \text{if}\; N_1 \geq\frac{1}{c_2} \log(c_1/\beta),\;n_y<2\\ \Big[ \frac{\log(c_1 / \beta)}{c_2 N_1} \Big]^{1/2} & \text{if}\; N_1 \geq\frac{1}{c_2} \log(c_1/\beta),\;n_y>2\\ \bar{\theta} & \text{if}\; N_1 \geq\frac{(\log 3)^2}{2} \log(c_1/\beta),\;n_y=2\end{cases}
\]
for $\bar{\theta}$ satisfying the condition
 \[
 \frac{\bar{\theta}}{\log (2+1/\bar{\theta})}=\bigg[\frac{\log (c_1/\beta)}{c_2 N_1}\bigg]^{1/2}.
 \]
Then, by the measure concentration inequality~\eqref{measure}, we have
 \[
 \mu^{N_1} \big \{
 \hat{w} \mid W (\mu, \nu) \leq \theta 
 \big \} \geq 1-\beta.
 \]
 It follows that for each $t$
 \begin{equation} \nonumber
 \begin{split}
 &\mu^{N_1} \Big \{
 \cvar_\alpha^{\mu} [ \dist (y^\star(t+1), \mathcal{Y}(t) + w_{\ell,t,1}) ] 
 \leq \sup_{\mu' \in \mathbb{D}} \cvar_\alpha^{\mu'} [ \dist (y^\star(t+1), \mathcal{Y}(t) + w_{\ell,t,1}) ] 
 \Big \} \geq 1-\beta.
 \end{split}
 \end{equation}
Since $\sup_{\mu' \in \mathbb{D}} \cvar_\alpha^{\mu'} [ \dist (y^\star(t+1), \mathcal{Y}(t) + w_{\ell,t,1}) ] \leq \delta_\ell$ by the definition of $y^\star$, we conclude that 
 the probabilistic performance guarantee~\eqref{Rel} holds with the choice of $\theta$ above. 
Similar results are also derived in~\cite[Theorem 3.5]{Esfahani2018} and 
~\cite[Theorem 3]{Yang2018}.
The constants $c_1$ and $c_2$ can be explicitly found using the proof of \cite[Theorem 2]{Fournier2015}. However, this choice often leads to an overly conservative radius $\theta$.
 One can obtain a less conservative $\theta$ by using bootstrapping or cross-validation methods~\cite{Esfahani2018}.
In the following section, we show how $\theta$ can be selected based on numerical experiments, depending on the choice of sample size.

\section{Numerical Experiments}\label{sec:sim}

In this section, we present simulation results that demonstrate the performance and the utility of the Wasserstein distributionally robust motion control method. We discuss two scenarios: $(i)$ a 5-dimensional nonlinear car-like vehicle model, and $(ii)$ a 12-dimensional linearized quadrotor model. 
For both cases, 
a reference trajectory is first generated with RRT* considering the dynamics of the vehicles. 
Then, the robotic vehicle starts following the trajectory while avoiding the randomly moving obstacles by solving the DR-MPC problem. We compare the performance of each robot for different sample sizes and Wasserstein radii. Moreover, we show the experimental results, demonstrating the advantage of using DR-MPC over the SAA-based risk-aware MPC method~\cite{Hakobyan2019}.
The parameters of the robot models in Table~\ref{Table1} were used throughout the simulations.
All the simulations were conducted on a PC with 3.70 GHz Intel Core i7-8700K processor and 32 GB RAM.

\begin{table}\centering\caption{Robotic vehicle parameters.}
\begin{tabular}{cc c cc}
 \hline
 \multicolumn{2}{c}{Car-like model} &&   \multicolumn{2}{c}{Quadrotor model} \\
 \cline{1-2} \cline{4-5}
 $m_V$ & $1700\; kg$ && $m_Q$ & $0.65\;kg$\\ 
 $C_f$ & $50\; kN/rad$  && $g$ & $9.81\;ms^2$\\ 
 $C_r$ & $50\; kN/rad$  && $l_Q$ & $0.23\;m$\\ 
 $I_z$ & $6000\; kg\cdot m^2$ && $I_{xx}$ & $0.0075\; kg\cdot m^2$\\ 
 $L_f$ & $1.2\; m$ && $I_{yy}$ & $0.0075\; kg\cdot m^2$\\ 
 $L_r$ & $1.3\; m$ && $I_{zz}$ & $0.013\; kg\cdot m^2$\\
 $v_x$ & $5\;m/s$  && &\\ 
 \hline
\end{tabular}\label{Table1}
\end{table}

\subsection{Nonlinear Car-Like Vehicle Model}

\begin{figure*}[t]
\centering
\begin{subfigure}{0.32\linewidth}
  \centering
  \includegraphics[width=\linewidth,trim={0.3in 0.1in 0.35in 0.25in},clip]{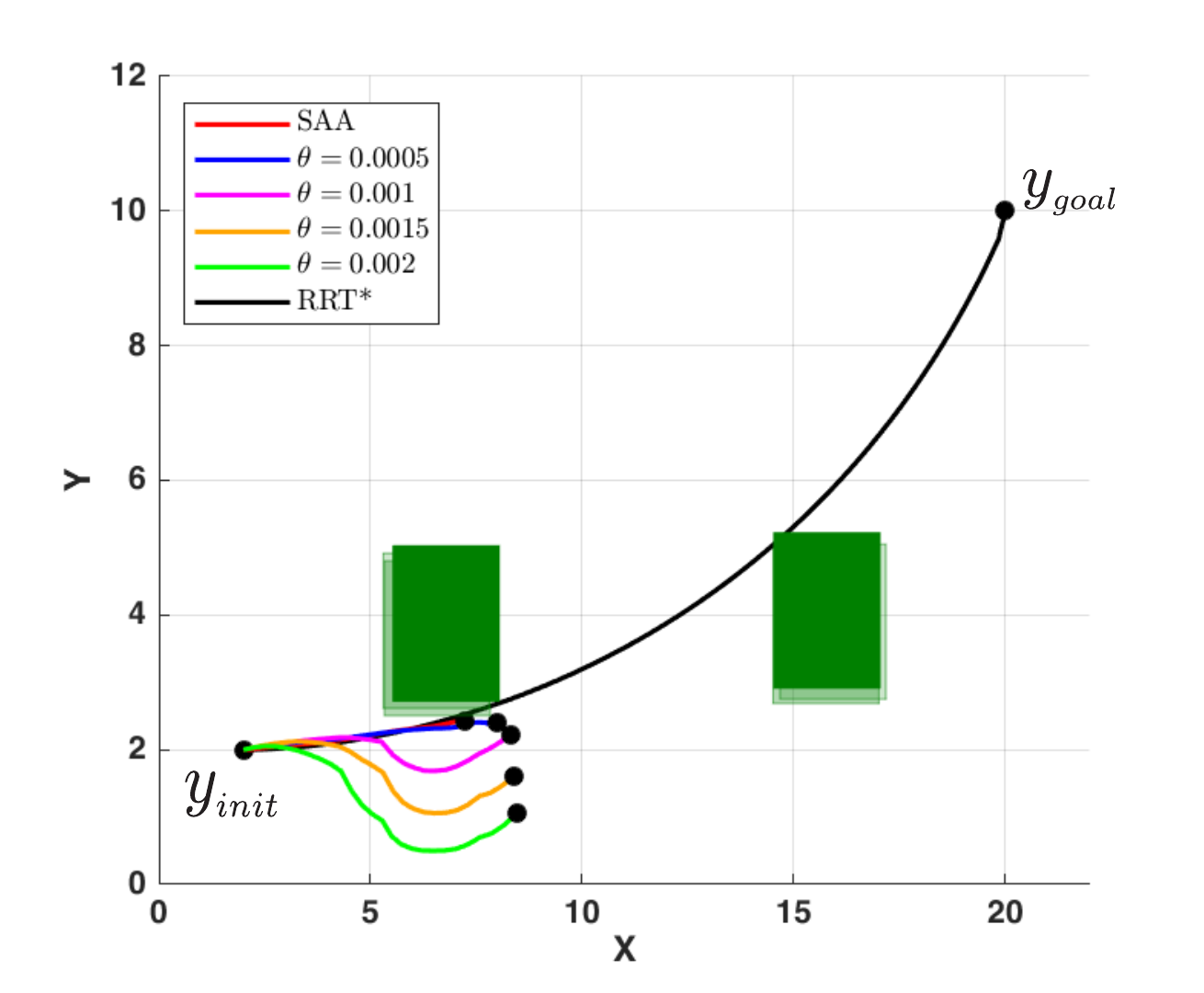}
  \caption{$t=26$}
\end{subfigure}%
\begin{subfigure}{0.32\linewidth}
  \centering
  \includegraphics[width=\linewidth,trim={0.3in 0.1in 0.35in 0.25in},clip]{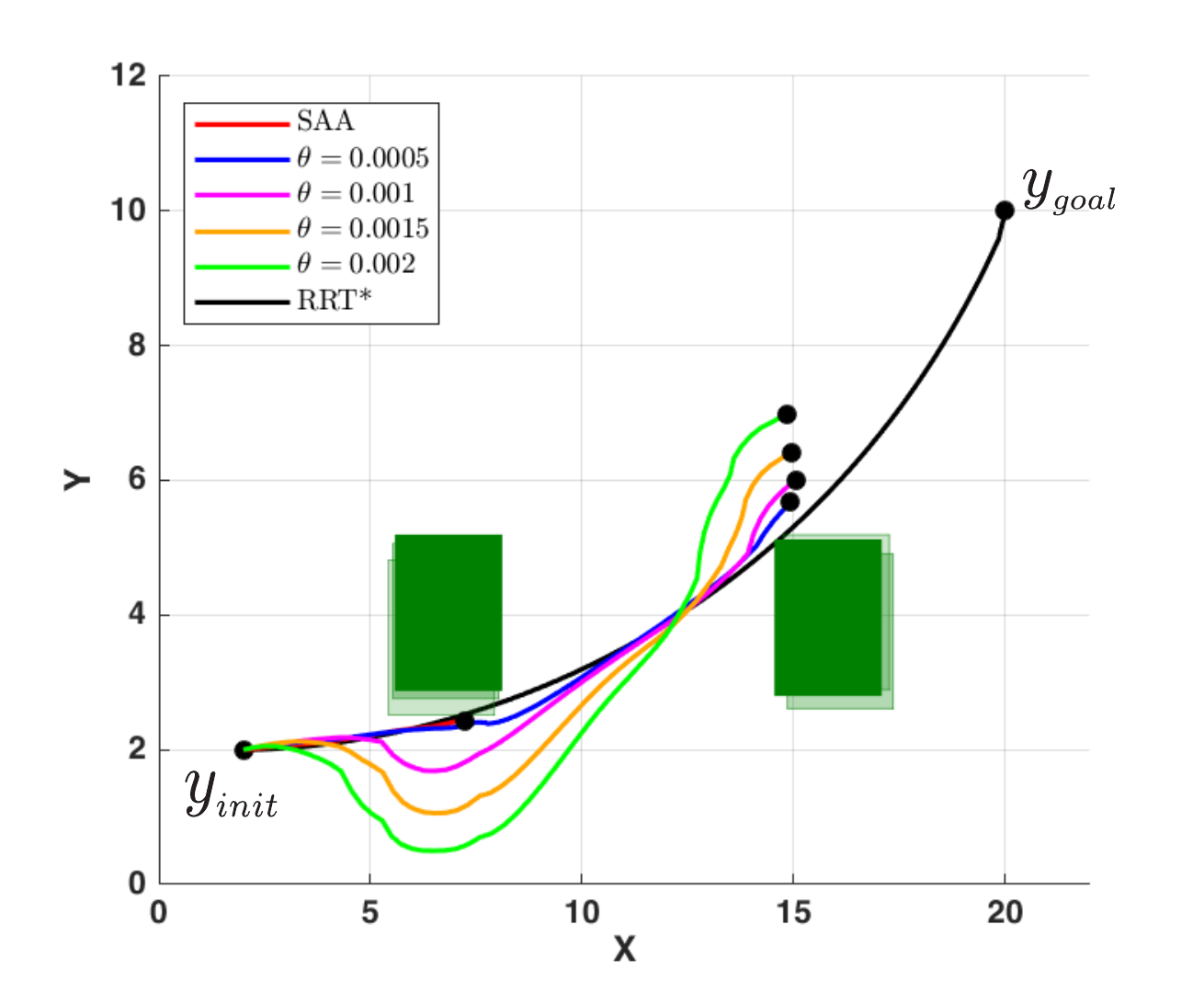}
  \caption{$t=56$}
\end{subfigure}%
\begin{subfigure}{0.32\linewidth}
  \centering
  \includegraphics[width=\linewidth,trim={0.3in 0.1in 0.35in 0.25in},clip]{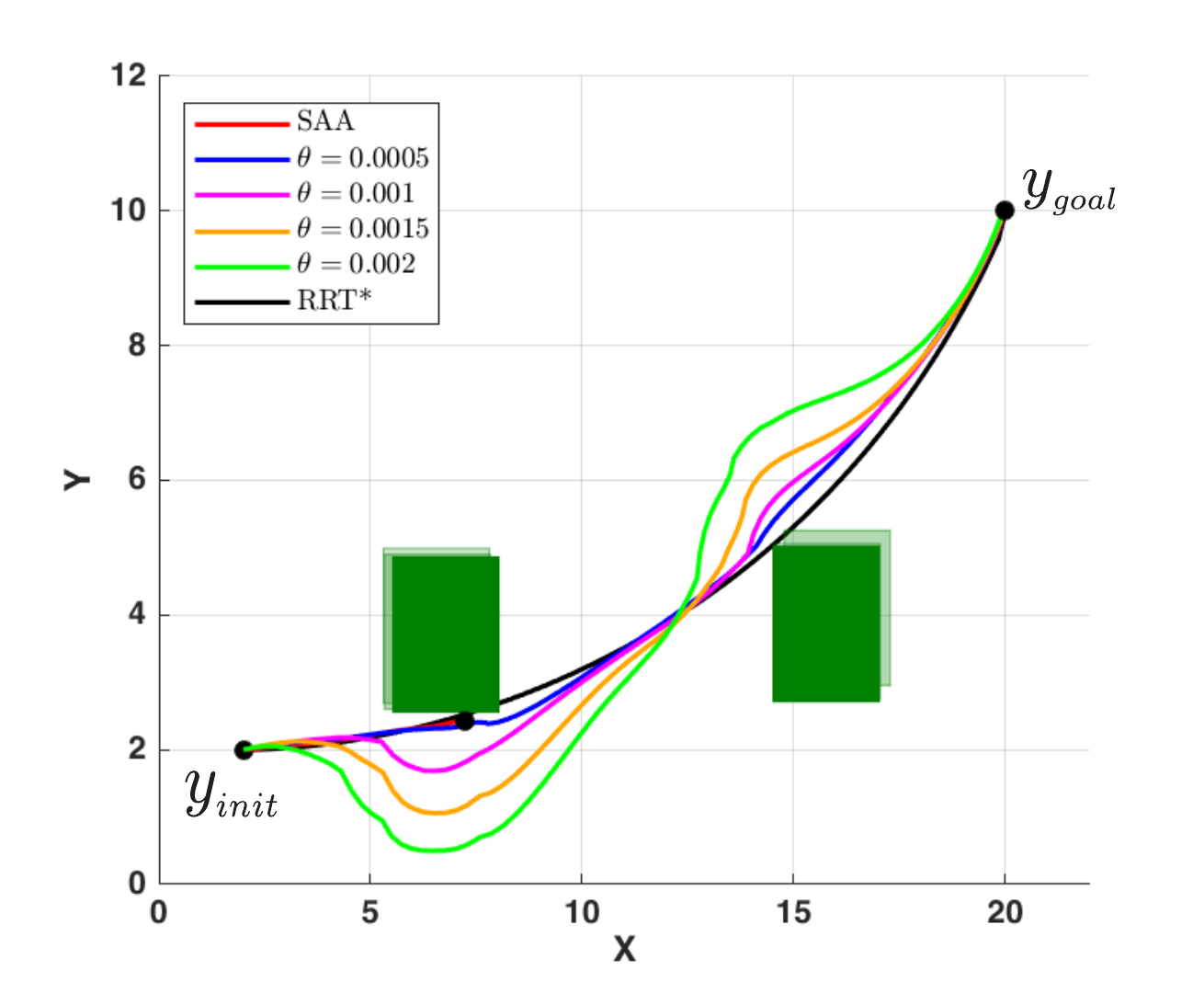}
  \caption{$t=80$}
\end{subfigure}%
\caption{Trajectories of the nonlinear car-like vehicle model controlled by SAA-MPC and DR-MPC with multiple $\theta$'s.}
\label{fig:Fig_Veh}
\end{figure*}

Consider a car-like vehicle navigating in a 2D environment with the following nonlinear model \cite{rajamani2011vehicle}:
\begin{align*}
&\dot{X}=v_x \cos\theta-v_y\sin\theta\\
&\dot{Y}=v_x\sin\theta+v_y\cos\theta\\
&\dot{\theta}=r\\
&\dot{v}_y=\frac{-2(C_f+C_r)}{m_V v_x} v_y-\Big(\frac{2l_fC_f-2l_rC_r}{m_V v_x}+v_x\Big)r+\frac{2C_f}{m_V}\delta_f\\
&\dot{r}=\frac{-2(l_fC_f+l_rC_r)}{I_z v_x} v_y-\frac{2l_f^2 C_f-2l_r^2 C_r}{I_z v_x}r+\frac{2l_fC_f}{I_z}\delta_f,
\end{align*}
where the state variables $X,Y,\theta,v_y$, and $r$ correspond to the vehicle's center of gravity in the inertial frame, lateral velocity, orientation and yaw rate, respectively. 
In addition, $v_x$ is the constant longitudinal velocity, $m_V$ is the mass of the vehicle, $I_z$ is the moment of inertia around the $z$ axis, $C_f$ and $C_r$ are the cornering stiffness coefficients for  the respective front and rear tires, and finally, $L_f$ and $L_r$ are the distances from the center of gravity to the front and rear wheels.
The output variables are chosen as the $X$ and $Y$ coordinates of the vehicle. 

The task is to design a controller that steers the vehicle to its goal position while avoiding the two randomly perturbing rectangular obstacles that are shown in Fig.~\ref{fig:Fig_Veh}. The random movement of each obstacle in each direction is sampled from a uniform distribution in $[-0.2,0.2]$. 
The MPC horizon is set to $K=20$. 
The weight matrix $Q$ is chosen as a $5 \times 5$ diagonal matrix with diagonal entries $(1,1,0,0,0)$. Let $P=1.2 Q$ and $R=0.01$. The MPC problem is solved for $T=80$ iterations using the discretized vehicle model with sample time $T_s=0.05\;sec$.
The interior-point method-based solver IPOPT was used to numerically solve the optimization problem~\eqref{WDR-MPC} at each MPC iteration.

\begin{table}[!t]
\caption{Computation time and operation cost for the nonlinear car-like vehicle motion control with $N_k=10$, $\delta_\ell=0.02$, and $\alpha=0.95$.}
\centering
\setlength{\tabcolsep}{0.5em} 
\begin{tabular}{>{\raggedright\arraybackslash}m{2.1cm}| c | c c c c}
\hline
&\multicolumn{1}{l | }{SAA-MPC} &   & \multicolumn{2}{c}{DR-MPC ($\theta$)}  &    \\
\cline{3-6}
 & & $0.0005$ & $0.001$ & $0.0015$ & $0.002$ \\
\hline\hline
\bf{Cost} & $+\infty$& 642.31 & 788.52 & 885.69 & 942.97 \\
\hline
\bf{Time (sec)}  & -&113.64 & 131.19 & 221.66 & 226.66 \\\hline
\end{tabular}
\label{Table2}
\end{table}

We first examine the effect of the Wasserstein ball radius~$\theta$ and compare DR-MPC with SAA-MPC~\cite{Hakobyan2019}.
Fig.~\ref{fig:Fig_Veh} shows the controlled trajectories for different $\theta$'s computed with $\delta_\ell=0.02$, $\alpha=0.95$ and $N_k=10$ sample data. 
As shown in Fig.~\ref{fig:Fig_Veh} (a),
in the early stages the vehicle follows the reference trajectory in the case of SAA-MPC and DR-MPC with small $\theta$, even though the robot gets closer to the first obstacle. 
However, in the case of DR-MPC with $\theta=0.0015$, the robot
proactively takes into account the obstacle's uncertainty for collision avoidance. 
The same behavior occurs for $\theta=0.002$ with a bigger safety margin. Thus, the robot further deviates from the reference trajectory. 
At $t=24$, the robot controlled by SAA-MPC violates the safety constraint and thus its operation is terminated. 
When DR-MPC is used, the robot passes the obstacle at $t=26$ without any collision.
The trajectory generated with $\theta =0.0005$ barely avoids the obstacle because the control action is not sufficiently robust. 
However, when a bigger radius is used, the robot avoids the obstacle with a wide enough safety margin. 
At $t=56$, the vehicle reaches the second obstacle. 
All four trajectories generated by DR-MPC are collision-free as desired. 
At  $t=80$, the vehicle reaches the goal position and the MPC iterations terminate.

Overall, 
we conclude that, with a small sample size $N_k = 10$, the SAA approach gives an infeasible result due to a violation of safety constraints, while the DR approach successfully avoids obstacles. 
Table \ref{Table2} shows the total computation time and the total cost $\sum_{t=0}^{T-1} r({x}^\star (t), {u}^\star (t))$ for SAA-MPC and DR-MPC with different $\theta$'s. 
The total cost increases with $\theta$ because a larger $\theta$ induces 
a more cautious control action that causes further deviations  from the reference path.
Thus, there is a fundamental tradeoff between   risk and cost.

\begin{figure*}[t]
\begin{subfigure}{0.5\linewidth}
\centering
\includegraphics[width=\linewidth]{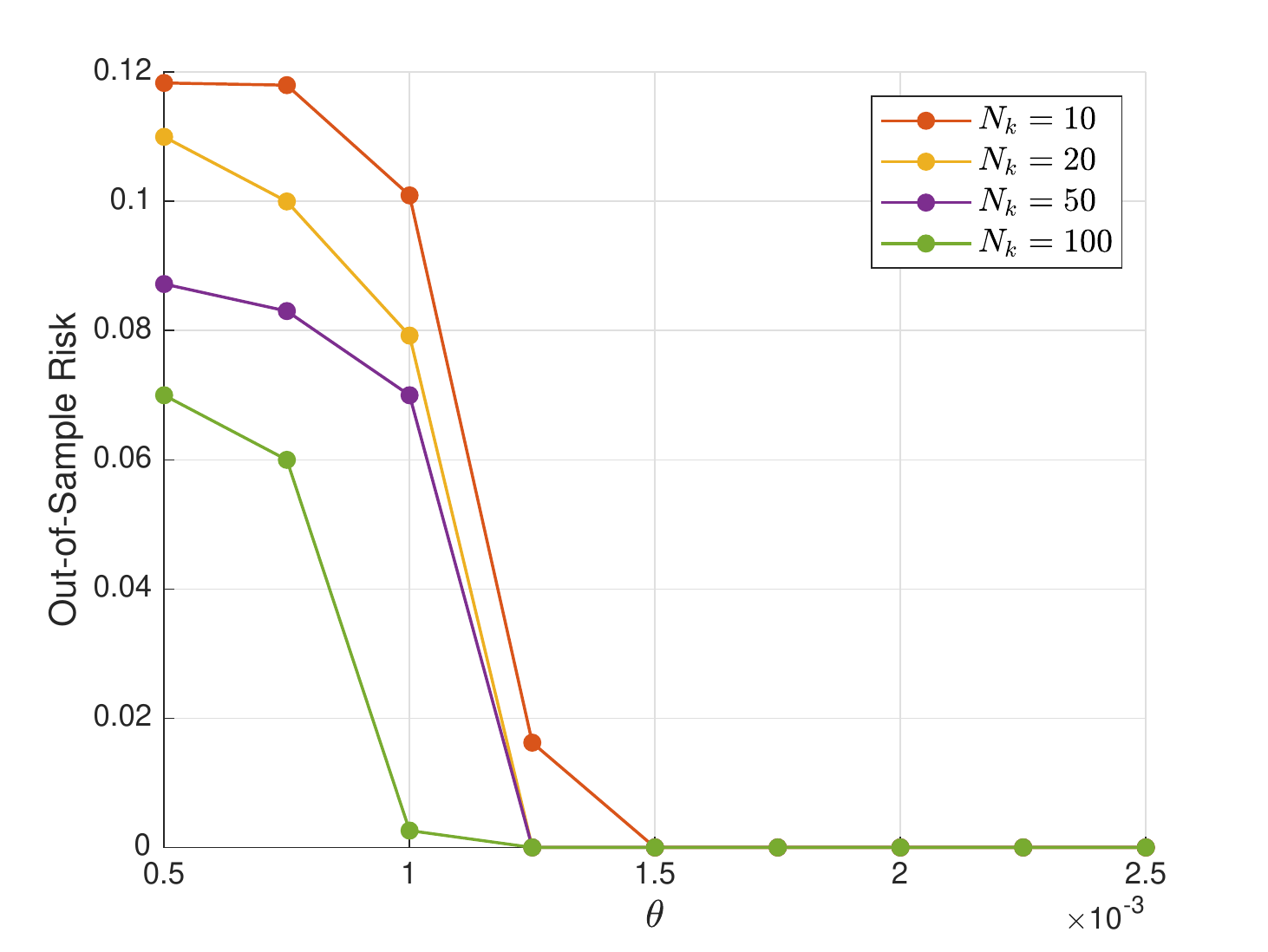}
\caption{ }
\label{fig:OSP_Veh}
\end{subfigure}%
\begin{subfigure}{0.5\linewidth}
\centering
\includegraphics[width=\linewidth]{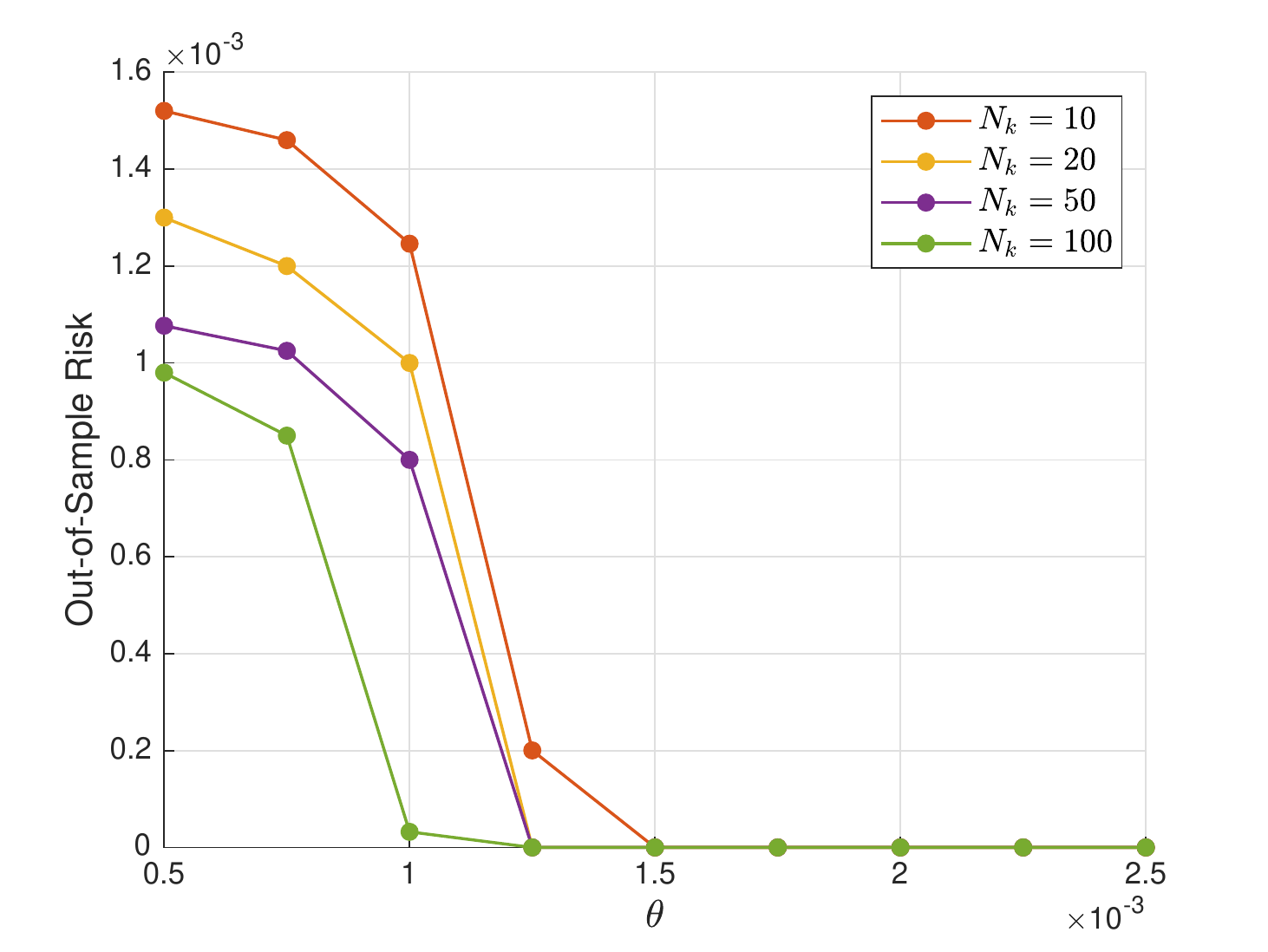}
\caption{ }
\label{fig:Avg_OPS_Veh}
\end{subfigure}
\caption{(a) Worst-case and (b) the average out-of-sample risk for the car-like vehicle.}
\end{figure*}

\begin{table}[t]
\caption{The worst-case reliability for the car-like vehicle motion control.}
\centering
\setlength{\tabcolsep}{0.5em} 
\begin{tabular}{>{\raggedright\arraybackslash}m{1cm}| c c c c c}
\hline
$N_k$ \textbackslash  $\theta$ & $0.0005$ & $0.00075$ & $0.001$ & $0.00125$ \\
\hline\hline
$10$ & $0.26$ & $0.26$ & $0.49$ & $1.00$\\
$20$ & $0.31$ & $0.66$ & $0.66$& $1.00$\\
$50$ & $0.55$ & $0.65$ & $0.74$ & $1.00$\\
$100$ & $0.60$ & $0.65$ & $1.00$ &$1.00$\\
\hline
\end{tabular}
\label{Rel_Veh}
\end{table}

We now investigate the out-of-sample safety risk by varying radius $\theta$ and sample size $N_k$. 
Specifically, for the $\ell$th obstacle we evaluate the worst-case out-of-sample risk
\[
\max_{t=0,\dots,T-1}\cvar_\alpha^\mu \big [\dist (y^\star(t+1), \mathcal{Y}(t) + w_{\ell,t,1}) \big],
\]
and the average out-of-sample risk
\[
\frac{1}{T}\sum_{t=0}^{T-1}\cvar_\alpha^\mu  \big [\dist (y^\star(t+1), \mathcal{Y}(t) + w_{\ell,t,1}) \big ].
\]
We estimated the CVaR using 20,000 independent samples generated from the true distribution $\mu$. 
The worst-case and average out-of-sample risks for different sample sizes and radii
are shown in Fig.~\ref{fig:OSP_Veh} and \ref{fig:Avg_OPS_Veh}, respectively. 
The worst-case out-of-sample risk is approximately 70 times larger than its average counterpart. 
Both out-of-sample risks monotonically decrease with radius $\theta$ and sample size $N_k$. 
Recall that the risk tolerance is chosen as $\delta_\ell = 0.02$.
In the case of $N_k = 10$, the risk constraints for all stages are satisfied if $\theta \geq 0.0015$. 
In all the other cases, the constraints are met for $\theta \geq 0.00125$.

For the probabilistic out-of-sample performance guarantee~\eqref{Rel},
we compute the worst-case reliability
\[
\min_{t=0,\dots,T-1}\mu^{N_1} \Big \{ \cvar_\alpha^{\mu}[ \dist (y^\star(t+1), \mathcal{Y}(t) + w_{\ell,t,1}) ]\leq \delta_\ell
\Big \}
\]
with 200 independent simulations with 1,000 samples in each.
Table~\ref{Rel_Veh} shows the estimated reliability depending on radius $\theta$ and sample size $N_k$. 
The reliability increases with $\theta$ and $N_k$ as expected. 
When $N_k=10$, the probability of meeting all the risk constraints for all stages is as low as $0.26$ (with a very small radius, $\theta = 0.0005$).
However, there is a sharp transition between $\theta = 0.001$ and $\theta = 0.00125$, and the reliability reaches its maximal value 1 when $\theta = 0.00125$.
In the case of larger sample sizes, e.g., $N_k = 100$, the reliability is relatively high even with a very small radius and reaches 1 when $\theta = 0.001$.

\subsection{Linearized Quadrotor Model}

Consider a quadrotor navigating in a 3D environment with the following linear dynamics:
\[ 
\begin{array}{lll}
\ddot{x} =-g\theta, & \ddot{y}=g\phi, & \ddot{z}=-\dfrac{l_Q}{m_Q}u_1, \\
\ddot{\phi}=\dfrac{1}{I_{xx}}u_2, & \ddot{\theta}=\dfrac{l_Q}{I_{yy}}u_3, & \ddot{\psi}=\dfrac{l_Q}{I_{zz}}u_4,
 \end{array}
\] 
where $m_Q$ is the quadrotor's mass, $g$ is the gravitational acceleration, and $I_{xx}$, $I_{yy}$ and $I_{zz}$ are the area moments of inertia about the principle axes in the body frame, and $l_Q$ represents the distance between the rotor and the center of mass of the quadrotor. The state of the quadrotor can be represented by its position and orientation with the corresponding velocities and rates in a 3D space---$(x,\dot{x},y,\dot{y},z,\dot{z},\phi,\dot{\phi},\theta,\dot{\theta},\psi,\dot{\psi}) \in \mathbb{R}^{12}$. The outputs are taken as the $X$, $Y$ and $Z$ coordinates of the quadrotor's center of mass.

\begin{figure*}[t]
\centering
\begin{subfigure}{0.45\linewidth}
  \centering
  \includegraphics[width=\linewidth,trim={0.25in 0.1in 0.2in 0.25in},clip]{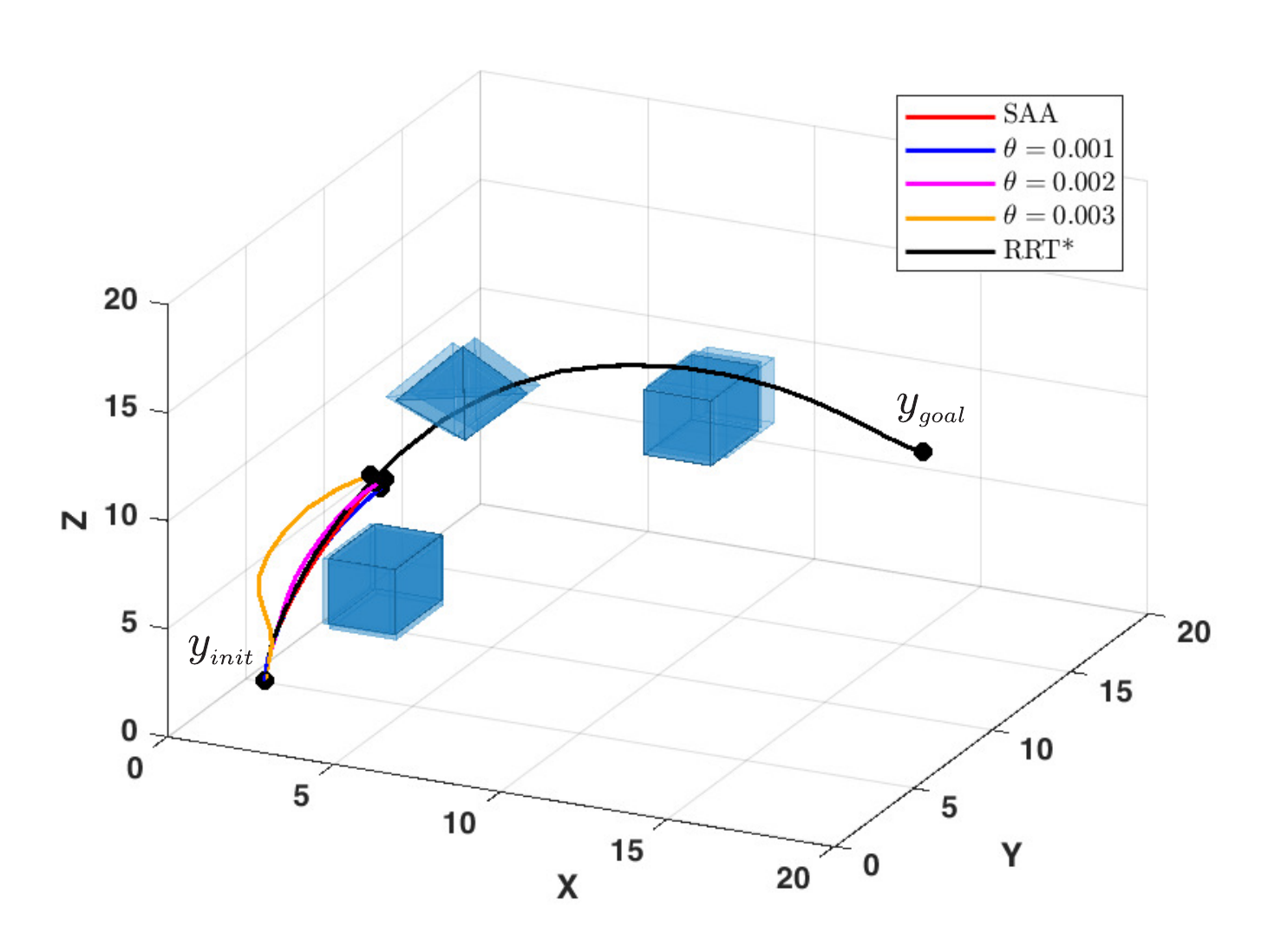}
  \caption{$t=15$}
\end{subfigure}%
\begin{subfigure}{0.45\linewidth}
  \centering
  \includegraphics[width=\linewidth,trim={0.25in 0.1in 0.2in 0.25in},clip]{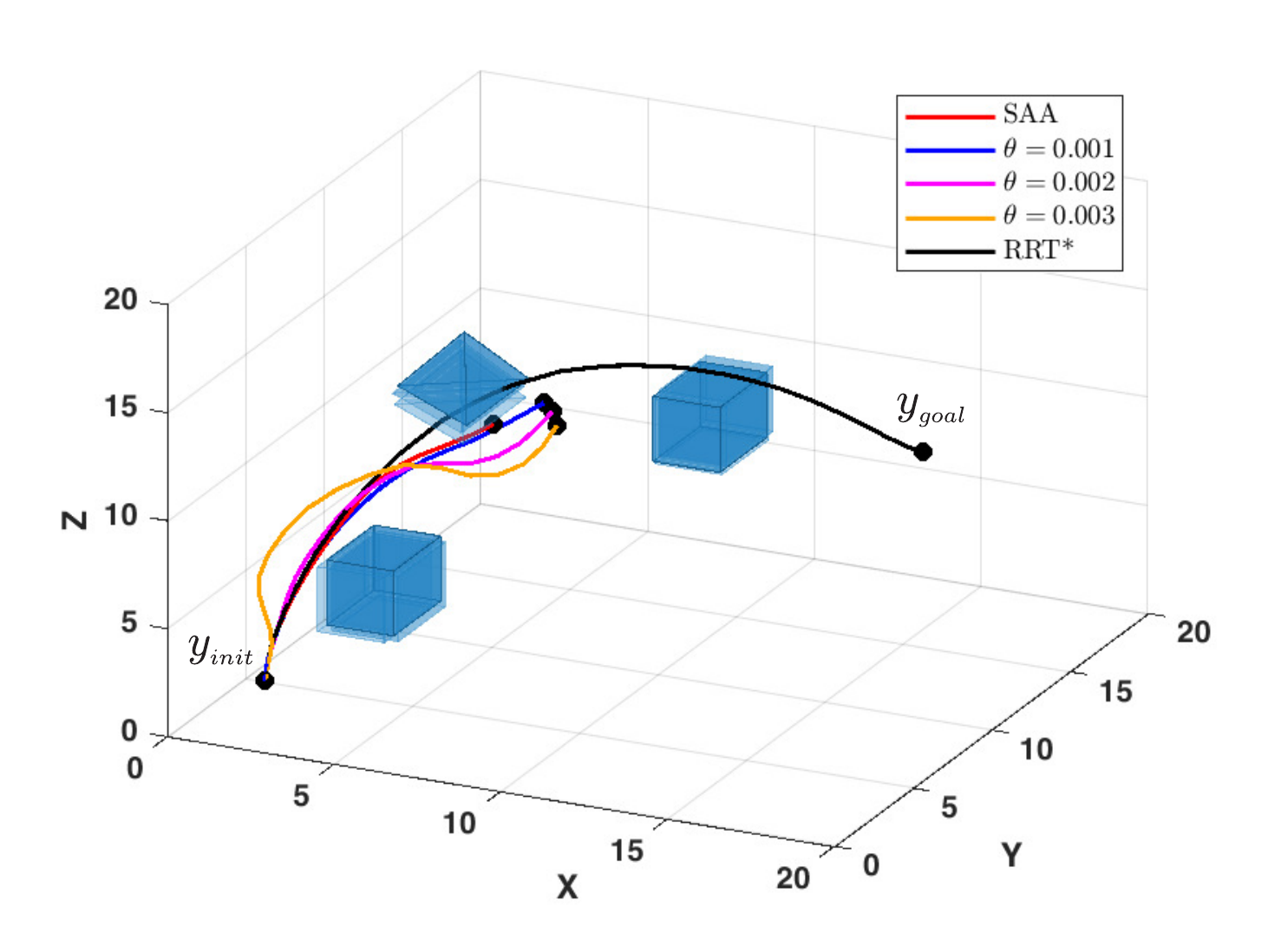}
  \caption{$t=22$}
\end{subfigure}
\begin{subfigure}{0.45\linewidth}
  \centering
  \includegraphics[width=\linewidth,trim={0.25in 0.1in 0.2in 0.25in},clip]{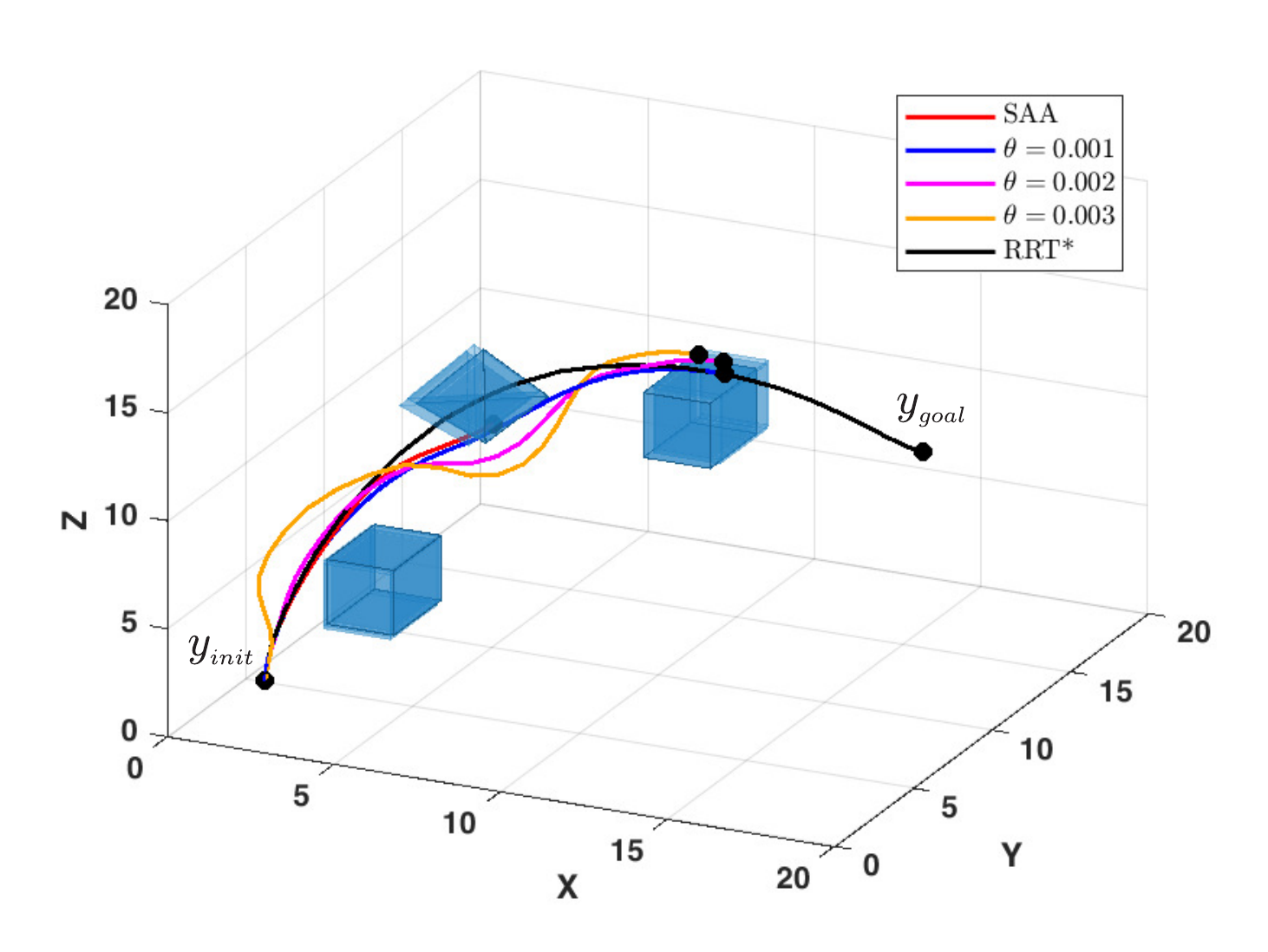}
  \caption{$t=30$}
\end{subfigure}%
\begin{subfigure}{0.45\linewidth}
  \centering
  \includegraphics[width=\linewidth,trim={0.25in 0.1in 0.2in 0.25in},clip]{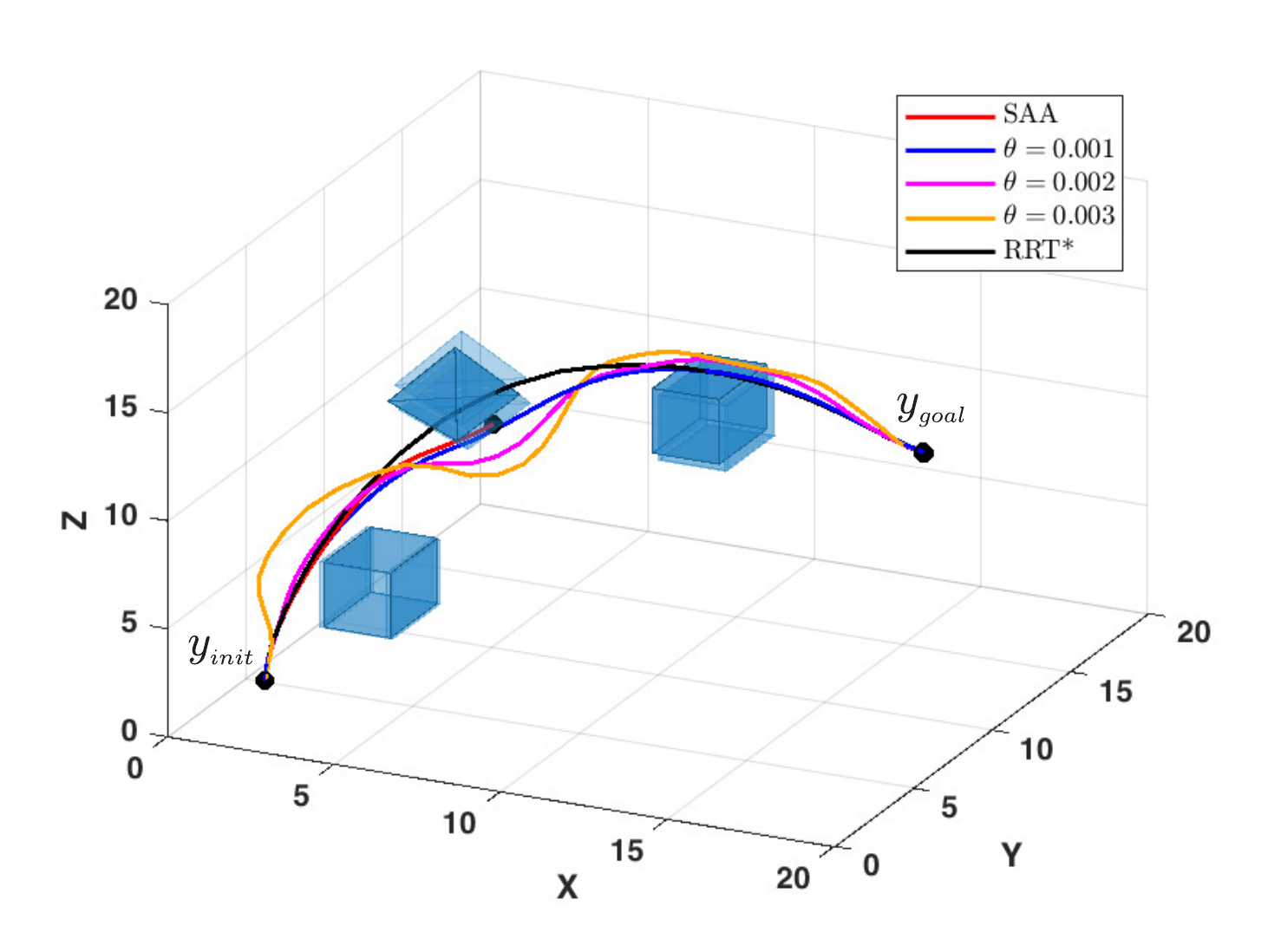}
  \caption{$t=50$}
\end{subfigure}
\caption{Trajectories of the quadrotor model controlled by SAA-MPC and DR-MPC with multiple $\theta$'s.}
\label{fig:Fig_Quad}
\end{figure*}

\begin{table}[t]
\caption{Computation time and operation cost for the quadrotor motion control with $N_k=10$, $\delta_\ell=0.02$, and $\alpha=0.95$.}
\centering
\setlength{\tabcolsep}{0.6em} 
\begin{tabular}{>{\raggedright\arraybackslash}c| c|| c | c c c}
\hline
\multicolumn{2}{r|}{\bf{Method}} & SAA & $\theta=0.001$ & $\theta=0.002$ & $\theta=0.003$ \\
\hline\hline
\multirow{2}{1.5cm}{\bf{Cost}} & \bf{NLP}& $+\infty$  & $14.02$ & $29.93$ & $86.34$ \\
&\bf{sBB} & $+\infty$ & $13.49$ & $28.86$ & $80.31$\\\cline{1-6}
\multirow{2}{1.5cm}{\bf{Time (sec)}} & \bf{NLP} & $-$ &  $47.69$ & $77.08$ & $136.02$\\
&\bf{sBB} & $-$ & $892.46$ & $6093.33$ & $9959.93$\\
\hline
\end{tabular}
\label{Table3}
\end{table}

The quadrotor is controlled to reach a desired goal position while avoiding three randomly perturbing obstacles. The random motions of the obstacles in each direction are drawn from the normal distributions $\mathcal{N}(0.2,0.1)$, $\mathcal{N}(-0.8,0.3)$ and $\mathcal{N}(0.3,0.2)$, respectively.
The MPC horizon is set to $K=10$.
The weight matrix $Q$ is selected as a $12 \times 12$ diagonal matrix with diagonal entries $(1, 0, 1, 0, 1, 0, 0, \ldots, 0)$.
We let $P = Q$ and $R = 0.02 I$. 
 The MPC problem is solved for $T=50$ iterations by discretizing the quadrotor model with sample time $T_s=0.1\;sec$.

The Wasserstein DR-MPC problem for the quadrotor model was solved using the sBB method with McCormick relaxation, as all the constraints are convex except the constraint \eqref{bl_const}. 
The relaxed problem in the algorithm was solved using the solver Gurobi, while the original one was solved using the solver IPOPT. 
In the initialization stage, $\underline{\varphi}$ and $\overline{\varphi}$ were chosen considering the quadrotor specifications. 
The bound on the control input was chosen based on the range of angular velocity of the rotors. 
Thus, the control input is restricted to the set $\mathcal{U}:=\{\bm{u} \in \mathbb{R}^4 \mid u_{\min}\leq \bm{u} \leq u_{\max}\} $ selected according to the motor specifications.\footnote{In the simulation, we used $u_{\min}=(0, -22.52, -22.52, -1.08)$ and $u_{\max}= (90, 22.52, 22.52,1.08)$.}
The state feasibility set $\mathcal{X}:=\{\bm{x}\in \mathbb{R}^{12} \mid -\pi \leq \phi \leq \pi\ ,\; -\frac{\pi}{2} \leq \theta \leq \frac{\pi}{2}, \; -\pi \leq \psi \leq \pi \}$ has been selected to limit the angles to avoid kinematic singularity.

The trajectories generated using $N_k=10$ samples with $\delta_\ell=0.02$ and $\alpha=0.95$ are shown in Fig.~\ref{fig:Fig_Quad}. 
We observe that for $t<15$ no collision occurs with the first obstacle. 
The trajectory with $\theta=0.003$ is the safest as its deviation from the reference trajectory is the largest. 
At $t=22$, the robot controlled by DR-MPC has passed the second moving obstacle while avoiding it.
However, 
in the case of SAA-MPC the safety constraint at $t=20$ is not satisfied, thereby resulting in a collision. 
At $t=30$, the quadrotor controlled by DR-MPC is near the third obstacle. Similar to the previous stages, trajectories with bigger $\theta$'s continue to deviate further from the risky reference trajectory with a larger operation cost as shown in Table~\ref{Table3}. At $t=50$, the robot completes the task and reaches the desired goal position.

\begin{figure*}[t]
\centering
\begin{subfigure}{0.5\linewidth}
\centering
\includegraphics[width=\linewidth]{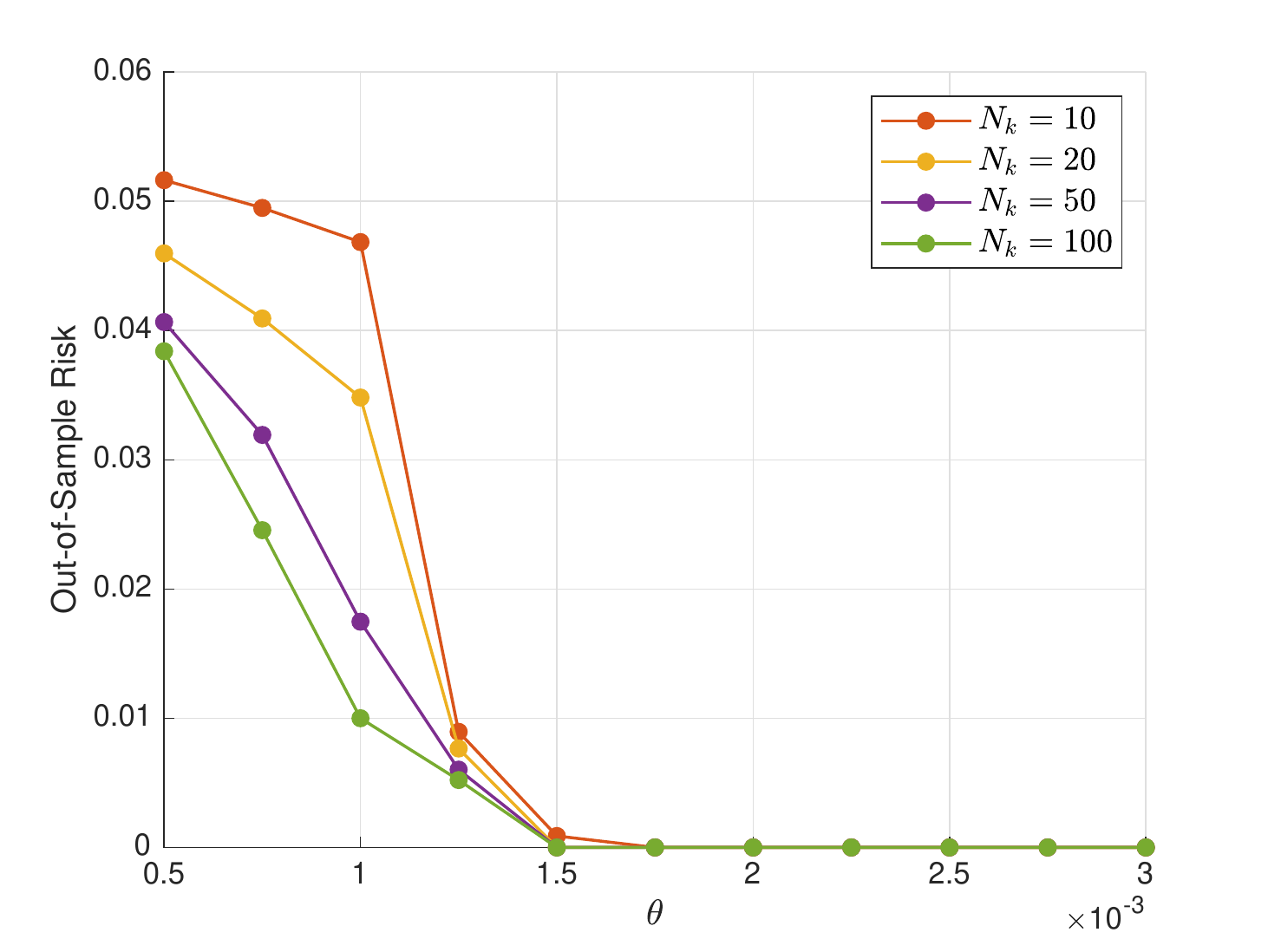}
\caption{ }
\label{fig:OSP_Quad}
\end{subfigure}%
\begin{subfigure}{0.5\linewidth}
\centering
\includegraphics[width=\linewidth]{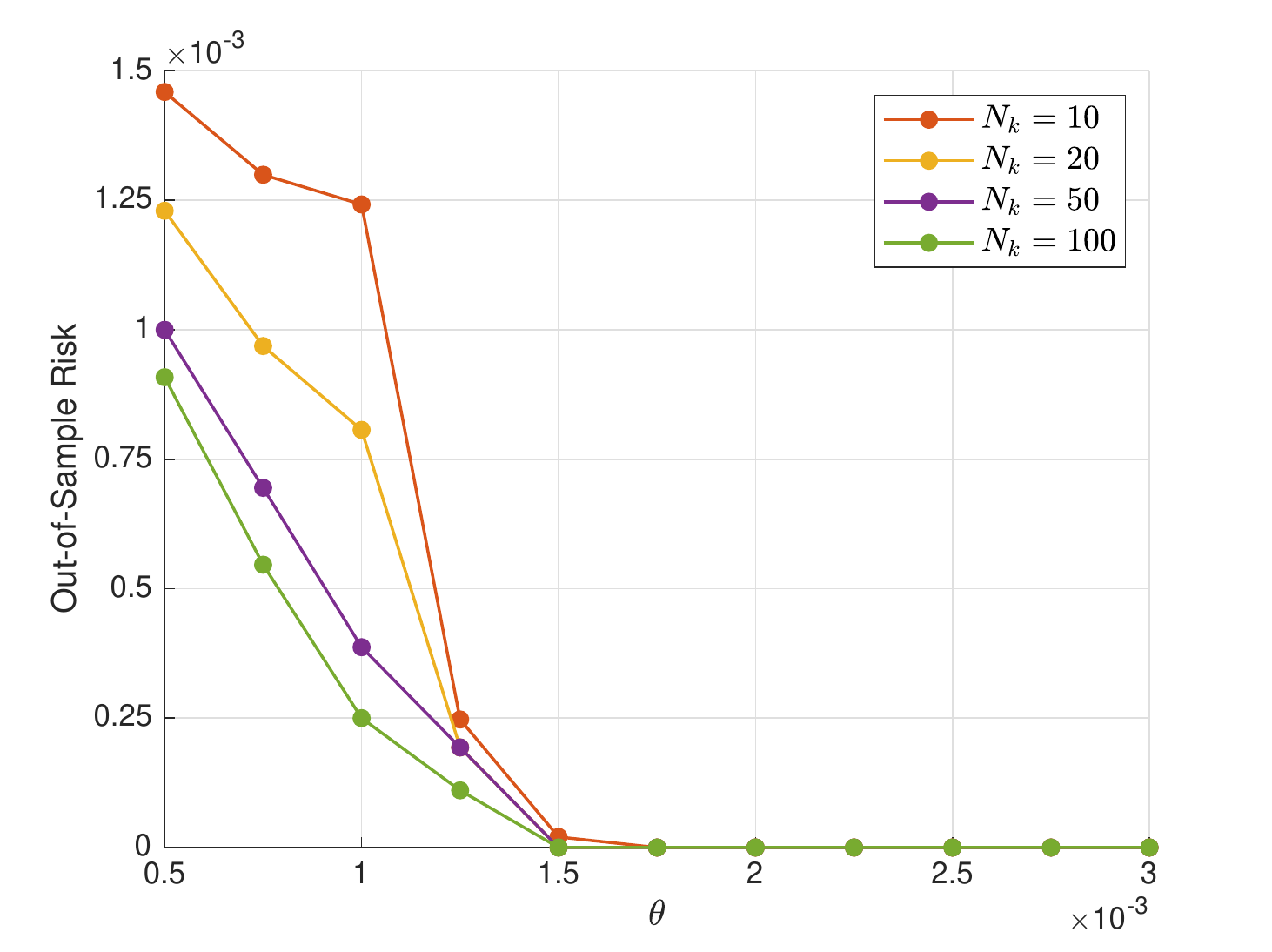}
\caption{ }
\label{fig:Avg_OSP_Quad}
\end{subfigure}
\caption{(a) Worst-case and (b) the average out-of-sample risk for the quadrotor.}
\end{figure*}

Table \ref{Table3} shows the computation time and the total cost $\sum_{t=0}^{T-1} r({x}^\star (t), {u}^\star (t))$ for SAA-MPC and DR-MPC with different $\theta$'s.
The Wasserstein DR-MPC problem is computed by two different methods: the sBB method with McCormick relaxation and the interior-point method implemented in IPOPT. 
Compared to SAA-MPC, 
Wasserstein DR-MPC shows a better performance in terms of the total cost and safety risk, while the computation time for SAA-MPC is lower than that for DR-MPC. 
From Table~\ref{Table3}, we observe that the cost obtained by sBB is less than that obtained by the interior-point method. 
This is consistent with the fact that sBB finds a globally optimal solution while the interior-point method converges to a local optimum. 
However, the interior-point method is faster than sBB as expected.

The selection of $\theta$ meeting the desired out-of-sample performance guarantee can be achieved by the same method as in the previous scenario. Figures \ref{fig:OSP_Quad} and \ref{fig:Avg_OSP_Quad} show the worst-case and  average out-of-sample risks estimated using 20,000 independent samples from the true distribution.
As expected, the out-of-sample risk decreases with the sample size and the ambiguity set size.
Table~\ref{Rel_Quad} shows the reliability $\min_{t=0,\dots,T-1}\mu^{N_1} \{\cvar[\dist(y_1,\mathcal{Y}(t)+w_{\ell,t,1})] \leq \delta_\ell\}$. The reliability does not significantly improve until $\theta=0.001$. Instead, it remains almost constant for all sample sizes when $\theta \leq 0.001$,
 and then rapidly increases. 
 A probabilistic guarantee of $0.92$ can be achieved on out-of-sample with only $20$ sample data and $\theta=0.00125$. 
 Thus, we can conclude that $0.00125$ is a reasonable choice for $\theta$ when only $20$ sample data are available, with which we achieve an acceptable out-of-sample performance guarantee.

\begin{table}[t]
\caption{The worst-case reliability for the quadrotor motion control.}
\centering
\setlength{\tabcolsep}{0.5em} 
\begin{tabular}{>{\raggedright\arraybackslash}m{1cm}| c c c c c}
\hline
$N_k$ \textbackslash  $\theta$ & $0.0005$ & $0.00075$ & $0.001$ & $0.00125$ & $0.0015$ \\
\hline\hline
$10$ & $0.54$ & $0.54$ & $0.64$ & $0.72$ & $1.00$\\
$20$ & $0.64$ & $0.64$ & $0.65$ & $0.92$ & $1.00$\\
$50$ & $0.65$ & $0.65$ & $0.69$ & $1.00$ & $1.00$\\
$100$ & $0.69$ & $0.69$ & $0.69$ & $1.00$ &$1.00$\\
\hline
\end{tabular}
\label{Rel_Quad}
\vspace{-0.1in}
\end{table}

\section{Conclusions}
In this work, we developed a risk-aware distributionally robust motion control method for avoiding collisions with randomly moving obstacles. 
By limiting the safety risk in the presence of distribution errors within a Wasserstein ball,
the proposed approach resolves the issue related to the inexact empirical distribution obtained from a small amount of available data and provides a probabilistic out-of-sample performance guarantee. 
The computational tractability of the resulting DR-MPC problem was achieved via a set of reformulations. 
Moreover, the sBB algorithm with McCormick relaxation was employed for obtaining a global optimal solution when the system dynamics and the output equations are affine. 
Finally, the performance of Wasserstein DR-MPC was demonstrated through numerical experiments on a nonlinear car-like vehicle model and a linearized quadrotor model. 
According to the simulation studies, even with a very small sample size $(N_k = 10)$,
Wasserstein DR-MPC successfully avoids randomly moving obstacles
and limits the out-of-sample safety risk (in a probabilistic manner),
 unlike the popular SAA method. 

The proposed distributionally robust motion control method can be extended in several interesting ways.
First, an explicit MPC method can be employed to reduce real-time computations.
Second, the proposed approach can be used in conjunction with Gaussian process regression to utilize the results of learning the obstacle's motion in a robust way. 
Third, a new relaxation method can  be developed for the case of nonlinear (polynomial) dynamics.

\bibliographystyle{IEEEtran}

\bibliography{reference}

\end{document}